\documentclass[lettersize,journal]{IEEEtran}
\usepackage{amsmath,amsfonts}
\usepackage{algorithmic}
\usepackage{array}
\usepackage{subcaption}

\usepackage{textcomp}
\usepackage{stfloats}
\usepackage{url}
\usepackage{verbatim}
\usepackage{graphicx}
\usepackage{cite}

\usepackage{amsmath,amssymb,amsfonts}
\usepackage{amssymb}
\usepackage{breqn}
\usepackage{cleveref}
\usepackage[lined,ruled,linesnumbered,commentsnumbered]{algorithm2e}
\usepackage[numbers,sort&compress]{natbib}
\usepackage{booktabs}
\usepackage{cases}
\usepackage{graphicx}
\usepackage{multirow}
\usepackage{mathrsfs}
\usepackage{textcomp}
\usepackage{xcolor}

\usepackage{bm}

\newtheorem{myThe}{Theorem}
\newtheorem{definition}{Definition}

\newcommand{\zz}[1]{{\textcolor{black}{ #1}}}
\begin{document}

\title{Multi-View Subgraph Neural Networks: Self-Supervised Learning with Scarce Labeled Data}

\author{
Zhenzhong Wang,
Qingyuan Zeng, 
Wanyu Lin,~\IEEEmembership{Member, IEEE},
Min Jiang,~\IEEEmembership{Senior Member, IEEE},
Kay Chen Tan,~\IEEEmembership{Fellow, IEEE}
\IEEEcompsocitemizethanks{\IEEEcompsocthanksitem

 Zhenzhong Wang, Wanyu Lin, and Kay Chen Tan are with the Department of Computing, The Hong Kong Polytechnic University, Hong Kong SAR, China (e-mail: zhenzhong16.wang@connect.polyu.hk; wan-yu.lin@polyu.edu.hk; kctan@polyu.edu.hk).
	
Min Jiang and Qingyuan Zeng are with the Department of Artificial Intelligence, Key Laboratory of Digital Protection and Intelligent Processing of Intangible Cultural Heritage of Fujian and Taiwan, Ministry of Culture and Tourism, School of Informatics, Xiamen University, Xiamen 361005, Fujian, China  (e-mail: minjiang@xmu.edu.cn;  36920221153145@stu.xmu.edu.cn).

\textit{Corresponding author: Wanyu Lin}


	}	
}

\markboth{Journal of \LaTeX\ Class Files,~Vol.~14, No.~8, August~2021}%
{Shell \MakeLowercase{\textit{et al.}}: A Sample Article Using IEEEtran.cls for IEEE Journals}


\maketitle

\begin{abstract}
While graph neural networks (GNNs) have become the {\em de-facto} standard for graph-based node classification, they impose a strong assumption on the availability of sufficient labeled samples. This assumption restricts the classification performance of prevailing GNNs on many real-world applications suffering from low-data regimes. Specifically, features extracted from scarce labeled nodes could not provide sufficient supervision for the unlabeled samples, leading to severe over-fitting. In this work, we point out that leveraging subgraphs to capture long-range dependencies can augment the representation of a node with homophily properties, thus alleviating the low-data regime. However, prior works leveraging subgraphs fail to capture the long-range dependencies among nodes. To this end, we present a novel self-supervised learning framework, called {\bf Mu}lti-view {\bf s}ubgraph n{\bf e}ural networks ({\em Muse}), for handling the long-range dependencies. In particular, we propose an information theory-based identification mechanism to identify two types of subgraphs from the views of input space and latent space, respectively. The former is to capture the local structure of the graph, while the latter captures the long-range dependencies among nodes. By fusing these two views of subgraphs, the learned representations can preserve the topological properties of the graph at large, including the local structure and long-range dependencies, thus maximizing their expressiveness for downstream node classification tasks. Theoretically, we provide the generalization error bound based on Rademacher complexity to show the effectiveness of capturing complementary information from subgraphs of multiple views. Empirically, we show a proof-of-concept of {\em Muse} on canonical node classification problems on graph data. Experimental results show that {\em Muse} outperforms the alternative methods on node classification tasks with limited labeled data. 
\end{abstract}


\section{Introduction}

\IEEEPARstart{G}{ragh} neural networks~\cite{10GraphConvolutionalNetworks,10GraphSAGE,GATvelivckovic2017graph} have been successful on a broad range of problems from diverse domains, e.g., social media analysis~\cite{wanyu-infocom20,Wang_Cao_Lin_Jiang_Tan_2023,ActionableExplanations}. Among others, several problems can be naturally cast as graph-based node classification tasks, including but not limited to artwork classification~\cite{GCNBoost}, video classification~\cite{9266105}, and content categorization~\cite{9868157,9431673}. For example, by analyzing the graph-based user interactions in a social network (\emph{e.g.}, Facebook, Twitter, Weibo), we can classify users, and then detect and recommend friends~\cite{cai2018comprehensive}. In essence, GNNs operate by a message-passing mechanism, where at each layer, nodes propagate their features to their neighbors. Being able to combine the topological information with feature information is what distinguishes GNNs from other purely topological learning approaches, such as label propagation~\cite{liu2019learning,li2018deeper,dong2021equivalence}, and arguably what leads to their success on graph-based node classification tasks.

\begin{figure}[!t]
  \centering
  \includegraphics[width=7.4cm]{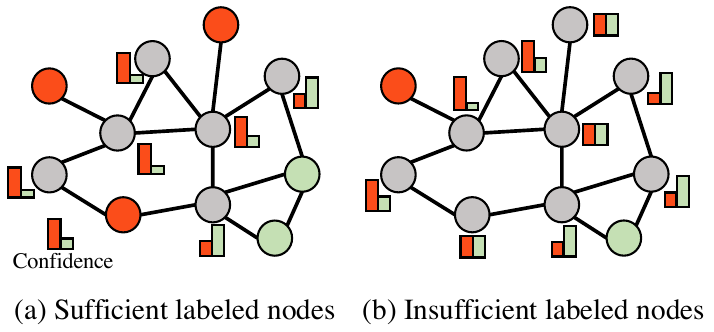}
  \caption{(a) Unlabeled nodes (in grey color) have high confidence to be predicted due to sufficient labeled nodes. (b) Unlabeled nodes are distant from the labeled nodes with homophily properties, thus having low confidence in predicting the class of the unlabeled nodes. }
  \label{fig:idea3}
\end{figure}

However, the performance of GNNs heavily relies on large amounts of labeled samples, hindering their applicability in many applications where labeled samples are extremely scarce and tricky to collect~\cite{101453442380112,chauhan2020fewshot,9157520}. When there are scarce labels in the graph, the unlabeled nodes can only obtain limited supervisory signals during the propagation process, leading to severe overfitting~\cite{9157520}. Specifically, GNNs enforce the embedding of two connected nodes to become similar by optimizing the Laplacian smoothing term~\cite{zhu2021interpreting}:
\begin{equation}\label{eq:tr}
\arg\min tr(\mathbf{H}^T\mathbf{L}\mathbf{H}),
\end{equation}
where $\mathbf{H}$ is the learned embedding, and $\mathbf{L}$ is the normalized symmetric positive semi-definite graph Laplacian matrix. This propagation mechanism will make a node's embedding similar to a labeled node with homophily properties. Naturally, it will have high confidence to be predicted as the label of this homomorphic node as long as the number of labeled nodes is sufficient (see Fig. \ref{fig:idea3} (a)). On the contrary, when only a few labeled samples are available, an unlabeled node may be distant from the labeled nodes with homophily properties, thus having low confidence in predicting the class of the unlabeled node (see Fig. \ref{fig:idea3} (b)).

\begin{figure}[!t]
  \centering
  \includegraphics[width=8.0cm]{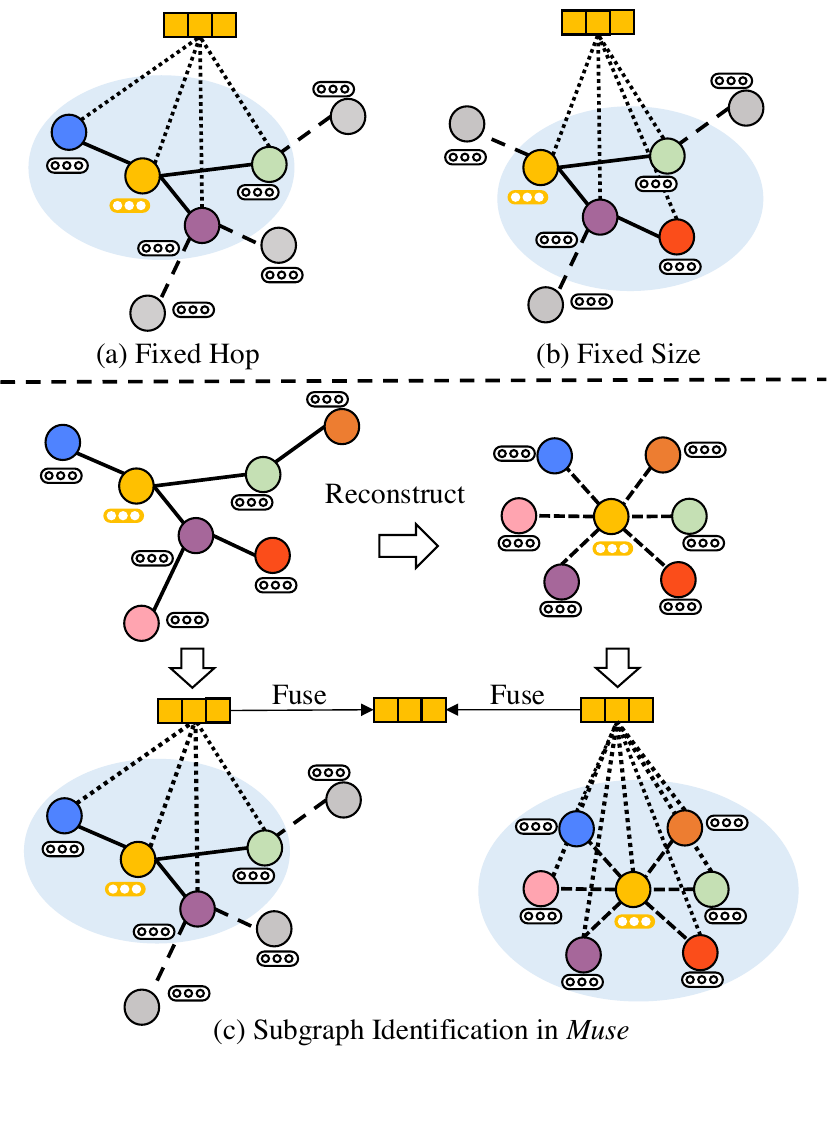}
  \caption{(a) The fixed hop limits the receptive field of the subgraph. (b) The structural information captured by the subgraph depends on the size of the subgraph. (c) Fusing two views of subgraphs can capture not only local structural information but also long-range dependencies.}
  \label{fig:idea}
\end{figure}

To alleviate the low-data dilemma, semi-supervised learning and self-supervised learning have emerged as promising paradigms. Semi-supervised learning constructs models using both labeled and unlabeled data~\cite{10192368}, while self-supervised learning (SSL) relies on pretext tasks constructed by unsupervised data to capture the supervision information~\cite{wu2024evolutionary,hao2023typing,zhai2019s4l,lee2020self}. Among SSL methods, context-based data augmentation has been proven to be a simple yet effective way that improve the generalization capability of the SSL models. However, most of these approaches are designed based on specific image transformation operators, \emph{e.g.}, rotation, cropping, and masking, and thus could not be applied to graph-structured data.

Recently, a few context-based SSL approaches have emerged for graph-based node classification in low-data regimes, each with its own perspective on this topic~\cite{9764632,liu2021self,sun2020multi,SelfSAGCNCVPR}. In particular, subgraphs have been regarded as an informative context that can augment the supervision signal in the context-based SSL on graphs. The intuition is that the nodes within a graph are interdependent; the local surroundings of nodes of interest, \emph{i.e.}, the structure of local subgraphs and corresponding neighbors' feature information, contain rich semantics that can be naturally used as supervision signals~\cite{xu2021self,jiao2020sub,graphicalmutualinformation}. However, the subgraphs defined by prior works are not flexible in the sense that the sizes or hops of subgraphs are predefined and fixed~\cite{Infomax,xu2021self,jiao2020sub,GmetaHuang,graphicalmutualinformation}. Specifically, the fixed number of hops tends to be small to prevent over-smoothing, while this limits the receptive field (see Fig.~\ref{fig:idea} (a)). On the other hand, if the predefined size of the subgraph is too large, irrelevant nodes will be involved; if the size is too small, limited relevant neighbors can be captured (see Fig.~\ref{fig:idea} (b)). Therefore, they may fail to capture the distant yet informative nodes, \emph{i.e.}, long-range dependencies that play essential roles in handling low-data regimes. That is, capturing long-range dependencies enables unlabeled nodes can perceive distant labeled nodes with homophily properties, thus improving the prediction confidence in Fig. \ref{fig:idea3} (b).


With the rich information of the subgraphs and the limitations of most existing subgraph-based SSL works in capturing long-range dependencies, this work proposes a new SSL approach based on multi-view subgraph neural networks, which can boost graph-based node classification performance under scarce labeled data. Specifically, we identify subgraphs from two different views: one view of the subgraphs comes from the original input space, and it can naturally capture the local structure for the labeled nodes. The other view of the subgraphs is extracted from the latent space, which is to capture the long-range dependencies of the nodes.  The rationality behind capturing the long-range dependencies from the latent space lies in the manifold assumption~\cite{4787647}, \emph{i.e.}, distant but similar data points are encouraged to map on a low-dimensional manifold in the latent space~\cite{pei2019geom}. Table \ref{tab:cos} empirically validates that complementary information can be captured by learning embedding from different spaces, where we calculate the mean cosine similarity of the same node's embedding extracted from three pairs of spaces: $O-L$ (the pair of the original input space $O$ and the latent space $L$), $O-O$, and $L-L$\footnote{The latent space is obtained by a classical manifold learning method Isomap~\cite{isomap}. A GNN is employed to extract the embedding from $O$ or $L$}. The smaller the value is, the more dissimilar the embeddings are, indicating different spaces indeed capture different information of nodes. Taking the cue, by fusing these two views of subgraphs (see Fig.\ref{fig:idea} (c)), we are able to capture the local structure and long-range dependencies of the labeled nodes within the graph, maximizing the expressiveness of learned representations with limited labeled nodes. The main contributions of our work are highlighted as follows,
\begin{enumerate}

\item We analyze that existing works on subgraph-based SSL fail to capture the long-range dependencies, leading to a sub-optimum performance for node classification tasks with limited labeled nodes. In addition, we provide a theoretical generalized error of the proposed {\em Muse} to illustrate the effectiveness of capturing complementary information from multiple views.

\item A novel multi-view subgraph-based SSL approach is proposed to capture both local structure and long-range dependencies of labeled nodes in the form of subgraphs.
For preserving the two topological properties at large, these subgraphs are fused as supervision signals for the downstream classification task. In particular, we propose a new information theory-based mechanism to identify the most related nodes of long-range dependencies by maximizing mutual information. Then, these related nodes are synthesized into subgraph representations to serve as supervision information.

\item We conduct a set of experiments on canonical node classification problems on graphs with benchmarking datasets. Our experimental results show that our method can achieve the best overall performance compared to alternative approaches based on supervised, semi-supervised, and SSL. In addition, various ablation studies and empirical analyses are conducted. We find that fusing subgraphs with different views can significantly improve the accuracy of node classification tasks in the low-data regime.

\end{enumerate}

The rest of this paper is organized as follows. Section II briefly reviews existing research about self-supervised learning and manifold learning. Section III details the designed {\em Muse}. In section IV, the experimental studies of the proposed algorithm and various state-of-the-art algorithms are presented. Finally, conclusions are drawn in Section V.

\begin{table}[htbp]
  \centering
  \caption{The mean cosine similarity between the embedding with respect to different pairs of spaces. The smaller the value is, the more dissimilar the embeddings are, indicating different spaces indeed capture different information of nodes.}
    \begin{tabular}{cccc}
    \toprule
    Dataset & $O-L$   & $O-O$   & $L-L$ \\
    \midrule
    Cora  & 0.2808 & 0.3008 & 0.2947 \\
    \midrule
    Citeseer & -0.3811 & -0.3995 & -0.3988 \\
    \midrule
    BlogCatalog & 0.5006 & 0.5093 & 0.5064 \\
    \bottomrule
    \end{tabular}%
  \label{tab:cos}%
\end{table}%

\begin{figure*}[h]
  \centering
  \includegraphics[width=0.78\linewidth]{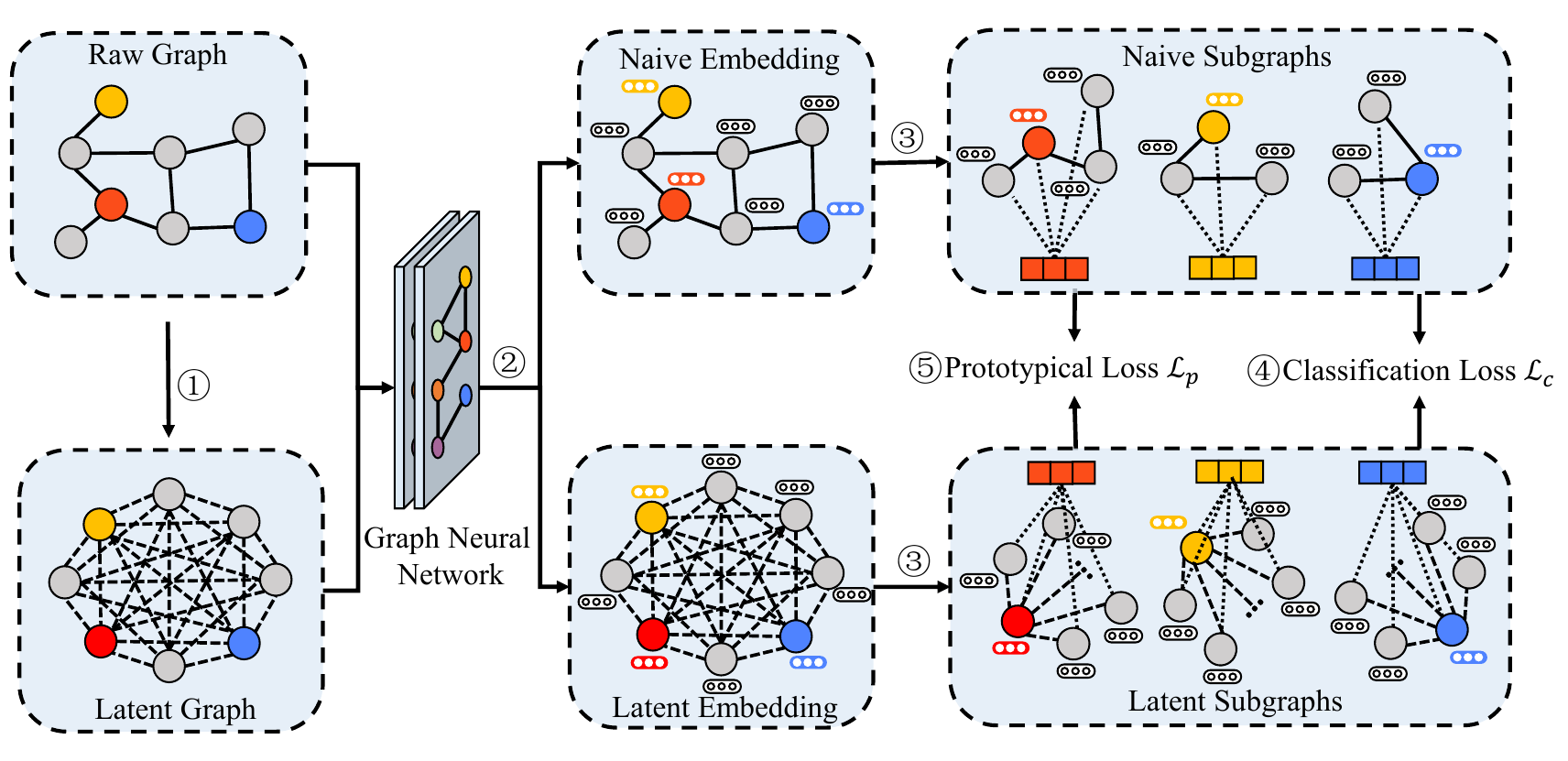}
  \caption{\textbf{Step 1}. The raw graph is reconstructed as a latent graph in which distant yet informative nodes can be mapped close, where grey nodes denote unlabeled nodes and colorful nodes denote labeled nodes.
  \textbf{Step 2}. A graph embedding network is employed to extract the naive embedding and the latent embedding from the raw graph and latent graph, respectively.
  \textbf{Step 3}. By maximizing mutual information, the naive subgraph and latent subgraph are respectively extracted from the naive embedding and the latent embedding.
  \textbf{Step 4}. Different embedding is fused together to achieve data augmentation, and the fused embedding is then used for calculating the classification loss. 
  \textbf{Step 5}. To leverage the inductive bias of different topological structures, a prototypical loss is derived by different subgraphs and node embedding.}
  \vspace{-10pt}
  \label{fig:pipeline}
\end{figure*}

\section{Related Work}

\subsection{Self-supervised
Learning }

SSL has recently emerged as a promising paradigm to overcome the challenge of lacking sufficient supervision. The key idea of SSL is to leverage the supervision information from a large amount of unlabeled data. In the field of computer vision, the images can be augmented by transformation operators (\emph{e.g.}, rotation, cropping, masking)~\cite{zhai2019s4l,lee2020self}. However, these approaches could not be applied to deal with graphs due to the inherent nature of graph-structured data. 


{\bf Graph-based SSL.} Recently, there have been a few works focusing on SSL in the domain of graphs. These methods can be roughly divided into two categories: context-based methods and contrastive-based methods. 
Context-based methods typically employ contextual information from a graph, \emph{e.g.}, various topological structures, to construct informative representations~\cite{xu2021self,sun2021sugar,graphicalmutualinformation}.
GraphLoG~\cite{xu2021self} hierarchically models both the local and global structure of a set of unlabeled graphs to infer graph-level representations. Sugar~\cite{sun2021sugar} and GMI~\cite{graphicalmutualinformation} learn discriminative representations by maximizing mutual information. Some works seek semantic information serving as supervised signals from nodes of an attributed graph ~\cite{SelfSAGCNCVPR,liu2021self}. M3S~\cite{sun2020multi} enlarges labeled set by assigning pseudo labels to unlabeled nodes with high confidence. Contrastive-based methods learn representations by measuring the metric distance between similar and dissimilar samples. For example, SUBG-CON~\cite{jiao2020sub} samples different subgraphs as positive and negative instances for learning. 
SelfSAGCN~\cite{SelfSAGCNCVPR} attempts to extract semantic information from nodes' features as supervised signals.  Similarly, SCRL~\cite{liu2021self} integrates topological information from both graph structure and node features as supervised information.

{\bf Subgraph-augmented Graph SSL.} The most related to ours is subgraph-augmented SSL on graphs. The local substructures, \emph{i.e.}, subgraphs in a graph contain vital features and prominent patterns, thus providing informative context for SSL. 
A handful of prior work has been devoted to mine subgraphs for graph representation learning~\cite{xu2021self,jiao2020sub,graphicalmutualinformation}. 
SUBG-CON~\cite{jiao2020sub} utilizes the strong correlation between central nodes and their regional subgraphs for inducing a contrastive loss. Sugar~\cite{sun2021sugar} and GMI~\cite{graphicalmutualinformation} utilize mutual information to measure the expressive ability of the obtained subgraph representations. GCC~\cite{qiu2020gcc} pretrains GNN for universe graph data by sampling two subgraphs for each node as a positive instance pair. CoLA~\cite{liu2021anomaly} captures the relationship between each node and its neighboring structure and uses an anomaly-related objective to train the contrastive learning model.


\zz{Long-range dependencies play a crucial role in graph representation learning. Taking the skeleton graph as an example, joints that are structurally apart can also have strong correlations~\cite{liu2020disentangling}. Existing subgraph-augmented graph methods often define subgraphs as local neighbors within a fixed size~\cite{sun2021sugar,jiao2020sub,Infomax} or fixed hop~\cite{graphicalmutualinformation,GmetaHuang,xu2021self}, they fail to capture the long-range dependencies, e.g., joints that are far apart but physically related to each other. In our work, we consider fusing subgraphs of two views: one kind of subgraph representation focuses on local information, and the other attempts to capture long-range dependencies to complement the representation of each other, thus maximizing the expressiveness of learned representations. }

\subsection{Multi-View Graph Learning}

\zz{Multi-view learning that leverages assorted types of features from heterogeneous views has promoted the performance of various machine learning tasks~\cite{fang2023efficient,zhang2023efficient,Wang_Cao_Lin_Jiang_Tan_2023,jiang2023adaptive}. Recently, a plethora of multi-view graph learning methods have been proposed concerning different downstream graph tasks. SelfSAGCN~\cite{SelfSAGCNCVPR} extracts semantic information from nodes' features as an additional view to enhance the original graph representation. To learn unbiased node representation, Graphair~\cite{ling2022learning} attempts to generate a fair view based on automated graph data augmentations as the complement for the biased view, thus mitigating the bias. Pro-MC~\cite{Wang_Cao_Lin_Jiang_Tan_2023} leverages noisy subgraph, smooth subgraph, and proxy subgraph as multi-view representation for learning a robust prior for graph meta-learning. EMSFS~\cite{zhang2023efficient} constructs a bipartite graph between training samples and generated anchors to complement the label propagation, thus facilitating the ultimate feature selection.}

\zz{The above works leverage various views such as noisy graphs, augmentation graphs, and semantic graphs as complementary information to enhance the representational ability, but they are not specifically designed for long-range dependencies. In this work, we fuse subgraphs from multiple views to capture both long-range dependencies and local information for low-data regimes.}

\subsection{Manifold Assumption}

Manifold learning aims to alleviate the curse of dimensionality, and it follows such an assumption: If high-dimensional data lie (roughly) on a low-dimensional manifold, then the data can be processed in a low-dimensional manifold space. More formally, the definition of the manifold assumption~\cite{4787647} is as follows: Suppose that the marginal probability distribution $P(x)$ underlying the data is supported on a low-dimensional manifold $\mathcal{M}$. Then the family of conditional distributions $P(y|x)$ is smooth, as a function of $x$, with respect to the underlying structure of the manifold $\mathcal{M}$.

Many real-world application data including images~\cite{102342342336,huang2020multimodal}, optimization~\cite{9552479,9097186,10292939}, and graph data~\cite{9155370,wang2014generalized,8027086} follows the manifold assumption. With the characteristic of manifold, learning algorithms can map high-dimensional data into a low-dimensional space and then essentially operate data in this low-dimensional space, thus avoiding the curse of dimensionality. 
During the past decades, a number of manifold learning techniques have been proposed. 
Locally Linear Embedding (LLE)~\cite{lle} focuses on sustaining the linear relationship between the sample points and their neighbors when projecting points into the low-dimensional space. 
Laplacian Eigenmaps (LE)~\cite{LaplacianEigenmaps} also pursues to reconstruct the local relationship between pairwise data. In LE, an adjacent matrix is used for representing the similarity between the data for reconstructing data.
Different from LLE and LE, Isomap~\cite{isomap} aims to retain the global geodesic distance of data in low-dimensional manifold space instead of only considering the local relationship. Generative adversarial neural networks~\cite{goodfellow2014generative} and the autoencoder models~\cite{vincent2008extracting} also demonstrate the promising ability to learn the intrinsic manifold of data. 
Their generator often learns the mapping from low-dimensional latent variables to the high-dimensional real data, and these low-dimensional latent variables can be seen as the embedding on the manifold.

\zz{In our work, the latent space of the data manifold is obtained using Isomap. Isomap transforms high-dimensional data to a lower dimension by a weighted graph. Initially, Isomap computes distances $d(i,j)$ between every pair of nodes $i$ and $j$ in the original high-dimensional feature space. Then, Isomap identifies which nodes are neighbors on the manifold $\mathcal{M}$ based on the distances $d(i,j)$ between pairs, representing these neighbors as a weighted graph $\mathcal{G}$ with edges of weight $d(i,j)$. Subsequently, Isomap defines the geodesic distance $d_{\mathcal{M}}(i,j)$ between all pairs by calculating the shortest path lengths by using the distances $d(i,j)$ in $\mathcal{G}$. As a result, a geodesic distance matrix $\mathbf{D}_{\mathcal{M}}$ is generated to indicate the shortest path lengths between all pairs of points in $\mathcal{G}$. Finally, by applying classical multidimensional scaling ~\cite{torgerson1952multidimensional} to the geodesic distance matrix $\mathbf{D}_{\mathcal{M}}$, Isomap creates a lower-dimensional embedding. The top $n$ eigenvectors of the geodesic distance matrix signify the embedding in the lower-dimensional space. More details of Isomap can be found in this reference~\cite{isomap}.}

The intuition behind capturing the long-range dependencies comes from the manifold assumption. Essentially, the manifold assumption indicates that similar data points can have a smaller distance in the low-dimensional space than that in the high-dimensional space. Taking this cue, our work aims to extract long-range dependencies from a low-dimensional latent space as complementary information.

\section{Proposed Algorithm}

Before introducing the proposed multi-view subgraph neural network, the problem setup of the limited labeled node classification is briefly given as follows, a graph $\mathcal{G}=(\mathbf{V},\mathbf{A},\mathbf{X})$ with a set of nodes $\mathbf{V}$ and an adjacency matrix $\mathbf{A}$ representing the connections is given, where $\mathbf{X}=(\mathbf{x_1},...,\mathbf{x_{|\mathbf{V}|}})^{T}\in \mathbb{R}^{|\mathbf{V}|\times d}$ is a set of feature vectors regarding nodes.
For limited labeled node classification, a set of labeled nodes $\mathbf{V}_l\subset \mathbf{V}$ with class labels from $\mathbf{Y}=\{y_1,...,y_K\}$ and a set of unlabeled nodes $\mathbf{V}_u\subset \mathbf{V}/\mathbf{V}_l$ are given. 
In particular, the nodes in  $\mathbf{V}_l$ is sparsely labeled $|\mathbf{V}_l|\ll |\mathbf{V}_u|$, \emph{e.g.}, 1 or 2 labeled samples per class in $\mathbf{V}_l$.
The goal of node classification is to map each node in $\mathbf{V}$ to one class in $\mathbf{Y}$.

  The workflow of the proposed multi-view subgraph neural network, {\em Muse}, is illustrated in Fig. \ref{fig:pipeline}, which is composed of the following key steps: The top and bottom branches extract the naive embedding and latent embedding from the raw graph data and latent graph data, respectively. Then, by maximizing mutual information between the labeled node and its neighbors, the naive subgraph and latent subgraph are respectively extracted from the naive embedding and the latent embedding. After that, subgraphs and node embedding are fused together to achieve data augmentation for the classification task. Moreover, a prototypical loss is designed to leverage the inductive bias of different embedding.

\subsection{Node Representation Learning}

In this work, two kinds of node embedding are extracted from graphs of two views. 
The first one, called naive embedding, comes from the raw graph data  $\mathcal{G}=(\mathbf{V},\mathbf{A},\mathbf{X})$. 
The other one is latent embedding extracting from a latent graph $\mathcal{G}^{'}=(\mathbf{V},\mathbf{A}^{'},\mathbf{X})$, where the adjacency matrix $\mathbf{A}^{'}$ is reconstructed by the original $\mathbf{A}$.

The basic idea of constructing the latent graph is that distant yet informative nodes can be mapped close in a latent space~\cite{pei2019geom}. To this end, a manifold algorithm, Isomap~\cite{isomap}, is employed to map features of nodes $\mathbf{X}$ into a low-dimensional latent space to get features $\mathbf{X}^{'}\in\mathbb{R}^{|\mathbf{V}|\times d^{'}}$. After that, the dot product is used to describe the similarity, \emph{i.e.}, adjacency matrix $\mathbf{A}^{'}$ of node pairs,
\begin{equation}\label{eq:dynamics}
\mathbf{A}^{'}=\mathbf{softmax}\left(\frac{\mathbf{X}^{'}{\mathbf{X}^{'}}^T}{\sqrt{d^{'}}}\right)\in \mathbb{R}^{|\mathbf{V}|\times |\mathbf{V}|}.
\end{equation}
In this manner, the reconstructed latent graph $\mathcal{G}^{'}=(\mathbf{V},\mathbf{A}^{'},\mathbf{X})$ can describe the distance metric in the latent space, and node pairs of long-range dependencies could have a high degree of similarity.

Naive embedding $\mathbf{H}$ and latent embedding $\mathbf{U}$ are respectively extracted by one GCN but with different propagation matrices $\mathbf{A}$ and $\mathbf{A}^{'}$. The naive embedding $\mathbf{H}^{(l+1)}$ and latent embedding $\mathbf{U}^{(l+1)}$ at $(l+1)$-th layer are respectively  encoded by the following layer-wise propagation equations,
\begin{equation}\label{eq:gcn1}
\begin{split}
\mathbf{H}^{(l+1)}=\sigma\left(\mathbf{D}^{-\frac{1}{2}}\widetilde{\mathbf{A}}\mathbf{D}^{-\frac{1}{2}}\mathbf{H}^{(l)}\mathbf{W}^{(l+1)}\right),\\
\mathbf{U}^{(l+1)}=\sigma\left({\mathbf{D}^{'}}^{-1}\widetilde{\mathbf{A}}^{'}\mathbf{U}^{(l)}\mathbf{W}^{(l+1)}\right),
\end{split}
\end{equation}
where $\widetilde{\mathbf{A}}=\mathbf{A}+\mathbf{I}$, $\widetilde{\mathbf{A}}^{'}=\mathbf{A}^{'}+\mathbf{I}$, $\mathbf{I}$ is an identity matrix, $\mathbf{D}$ and $\mathbf{D}^{'}$ are diagonal matrices with $\mathbf{D}[i,i]=\sum_{j}\mathbf{A}[i,j]$ and $\mathbf{D}^{'}[i,i]=\sum_{j}\mathbf{A}^{'}[i,j]$, respectively, $\sigma(\cdot)$ denotes an activation function, $\mathbf{W}^{(l)}$ is a layer-specific trainable weights, and $\mathbf{H}^{(0)}=\mathbf{U}^{(0)}=\mathbf{X}$. 

For simplicity, the naive embedding and latent embedding of the last layer with respect to the inputs $\mathcal{G}$ and $\mathcal{G}^{'}$ are respectively denoted as $\mathbf{H}\in\mathbb{R}^{|\mathbf{V}|\times K}$ and $\mathbf{U}\in\mathbb{R}^{|\mathbf{V}|\times K}$.
 
\subsection{Self-supervised Multi-View Subgraph Augmentation}

For a given labeled node $i\in\mathbf{V}_l$, subgraph identification mines multi-view subgraphs, \emph{i.e.}, naive subgraph $\mathbf{S}_{i}^{\mathbf{H}}$ and latent subgraph $\mathbf{S}_{i}^{\mathbf{U}}$ from $\mathbf{H}$ and $\mathbf{\mathbf{U}}$, respectively, for augmenting the representation of the given node $i$ ($\mathbf{S}_{i}^{\mathbf{H}}\subset\mathbf{H}$, $\mathbf{S}_{i}^{\mathbf{U}}\subset\mathbf{\mathbf{U}}$).
The naive subgraph can capture local neighboring information since the design of prevailing GNNs is shallow to avoid over-smoothing~\cite{li2018deeper}. 
While the latent subgraph is constructed in a latent space enabling long-range dependencies to be captured. 
After identifying multi-view subgraphs, the representation of the labeled node $i$ is augmented by fusing $\mathbf{S}_{i}^{\mathbf{H}}$ and $\mathbf{S}_{i}^{\mathbf{U}}$.

To determine the more correlated node embedding to node $i$ to form $\mathbf{S}_{i}^{\mathbf{\Psi}}$ ($\mathbf{\Psi}$ denotes identifier the $\mathbf{H}$ or $\mathbf{U}$ for simplicity),
the subset $\mathbf{S}_{i}^{\mathbf{\Psi}}$ can be randomly sampled from $\mathbf{\Psi}$ by maximizing mutual information (MI),

\begin{equation}\label{eq:MI}
\max E_{\mathbf{S}_i^{\mathbf{\Psi}}\subset\mathbf{\Psi}}\left[MI(\mathbf{\Psi}_i,\mathbf{S}^{\mathbf{\Psi}}_i)\right]=E[H(\mathbf{\Psi}_i)]-E[H(\mathbf{\Psi}_i|\mathbf{S}^{\mathbf{\Psi}}_i)],
\end{equation}
where $\mathbf{\Psi}_i$ is the naive embedding or latent embedding of the labeled node $i$, and $H(\cdot)$ is the entropy term. In this way, the change of the correlation between the subgraph formed by different nodes and the embedding $\mathbf{\Psi}_i$ can be measured. When the GNN is trained and fixed in an epoch, $\mathbf{\Psi}_i$ is constant.
Therefore, we just need to minimize the upper bound of the second term in Eq. (\ref{eq:MI}) by applying Jensen’s inequality with the convexity assumption,
\begin{equation}\label{eq:conditionalentropy}
\min_{\mathbf{S}_i^{\mathbf{\Psi}}\subset\mathbf{\Psi}}  H\left(\mathbf{\Psi}_i|E[\mathbf{S}^{\mathbf{\Psi}}_i]\right).
\end{equation}
Although the convexity assumption cannot be satisfied due to the complexity of neural networks, minimizing the upper bound can also be an alternative method.


\zz{Because there are exponential combinations of $\mathbf{S}_i^{\mathbf{\Psi}}\subset\mathbf{\Psi}$, directly estimating $E[\mathbf{S}^{\mathbf{\Psi}}_i]$ in Eq. (\ref{eq:conditionalentropy}) is not tractable. To tractably estimate $P(\mathbf{S}^{\mathbf{\Psi}}_i)$, we transform the combinational problem of forming $\mathbf{S}_i^{\mathbf{\Psi}}$ as a multivariate Bernoulli distribution. Specifically, the probability of selecting $\mathbf{\Psi}_j\in \mathbf{\Psi} $ as one of the related node embedding for forming $\mathbf{S}_i^{\mathbf{\Psi}}$ is denoted as $P(\mathbf{\Psi}_j)$. Then, the probability of selecting all related node embedding over all $\mathbf{\Psi}_j\in \mathbf{\Psi}$ for forming $\mathbf{S}_i^{\mathbf{\Psi}}$ is a multivariate Bernoulli distribution,}


\begin{equation}\label{eq:Bernoulli}
P(\mathbf{S}^{\mathbf{\Psi}}_i)=\prod_{\mathbf{\Psi}_j\in \mathbf{\Psi}}P(\mathbf{\Psi}_j).
\end{equation}
\zz{The probability $P(\mathbf{\Psi}_j)$ can be represented by masking $\mathbf{\Psi}$ with a mask vector $\mathbf{M}^{\mathbf{\Psi}}_i \in \mathbb{R}^{|\mathbf{V}|\times 1}$, in which each entry $\mathbf{M}^{\mathbf{\Psi}}_i[j]$ represents the probability of existence of $\mathbf{\Psi}_j$ existing in $\mathbf{S}^{\mathbf{\Psi}}_i$.} By masking, the conditional entropy in Eq. (\ref{eq:conditionalentropy}) can be replaced with



\begin{equation}\label{eq:finalconditionalentropy}
\min_{\mathbf{M}^{\mathbf{\Psi}}_i}  H\left(\mathbf{\Psi}_i|\mathbf{S}^{\mathbf{\Psi}}_i=\sigma(\mathbf{M}^{\mathbf{\Psi}}_i) \odot \mathbf{\Psi}\right),
\end{equation}
where $\odot$ denotes element-wise multiplication, and $\sigma(\cdot)$ denotes the sigmoid function that maps the mask entry to $[0,1]$. For computational efficiency, the entry of nodes beyond $k$-hop ($k=3$) in the mask $\mathbf{M_i^H}$ is set to 0, and
the $j$-th entry in the mask $\mathbf{M_i^U}[j]$ is set to 0, where $\mathbf{A}^{'}[i,j]<\tau$ ($\tau=0.5$). The embedding of nodes whose entries are equal to 0 will not appear in the $\mathbf{S}^{\mathbf{\Psi}}_i$.

Furthermore, when the network is trained and then fixed, we can approximate the conditional entropy objective in Eq.  (\ref{eq:finalconditionalentropy}) with the Kullback–Leibler divergence between the subgraph $\mathbf{S}_i^{\mathbf{\Psi}}$ and the embedding $\mathbf{\Psi}_i$ of the labeled node $i$, so that $\mathbf{M}_i^{\Psi}$ can be optimized by a few steps of gradient descent as follows,
\begin{equation}\label{eq:finalentropy}
\min_{\mathbf{M}^{\mathbf{\Psi}}_i} KL(\mathbf{\Psi}_{i}||\mathbf{S}^{\mathbf{\Psi}}_{i}) =\sum_{k}\mathbf{\Psi}_{i,k}\log \frac{\mathbf{\Psi}_{i,k}}{\sum_{\mathbf{\Psi}_j\in\mathbf{S}^{\mathbf{\Psi}}_{i}}\mathbf{\Psi}_{j,k}},
\end{equation}
where $\mathbf{\Psi}_{i,k}$ is the $k$-th element of $\mathbf{\Psi}_{i}$, and $\mathbf{S}^{\mathbf{\Psi}}_i= \sigma(\mathbf{M}^{\mathbf{\Psi}}_i) \odot \mathbf{\Psi}$.

Each entry in $\mathbf{M}_i^{\Psi}$ indicates the degree of correlation of the corresponding node to the labeled node $i$. For example, a higher $\mathbf{M}_i^{\Psi}[j]$ means the node $j$ is more correlated to node $i$, and we can use node $j$'s embedding containing more correlation to enrich $i$'s representation. Therefore, we regard each entry in $\sigma(\mathbf{M}_i^{\Psi})$ as the weight of the node to form the subgraph $\mathbf{S}^{\mathbf{\Psi}}_i$, and subgraph-level embedding $\widetilde{\mathbf{S}}_{i}^{\Psi}\in\mathbb{R}^{K}$ can be  written as the weighted average of all embedding in this subgraph,
\begin{equation}\label{eq:subgraphs}
     \widetilde{\mathbf{S}}_{i}^{\Psi}=\frac{\sum_{j} \mathbf{\Psi}_j \sigma\left(\mathbf{M}^{\mathbf{\Psi}}_i[j]\right)}{\sum_{j}\sigma\left(\mathbf{M}^{\mathbf{\Psi}}_i[j]\right)}, \mathbf{\Psi}_j\in \mathbf{S}^{\mathbf{\Psi}}_{i}.
\end{equation}

Finally, we fuse naive subgraph $\widetilde{\mathbf{S}}_{i}^{\mathbf{H}}$, latent subgraph $\widetilde{\mathbf{S}}_{i}^{\mathbf{U}}$, latent embedding $\mathbf{U}_i$, and naive embedding $\mathbf{H}_i$ together. The concatenation embedding is inputted into a single-layer fully connected network $FC(\cdot)$ as the final augmentation representation $\overline{\mathbf{H}}_i$ for the node $i$, 
\begin{equation}\label{eq:finalembedding}
\overline{\mathbf{H}}_i=\sigma\left(FC(\widetilde{\mathbf{S}}_{i}^{\mathbf{H}} \oplus \widetilde{\mathbf{S}}_{i}^{\mathbf{U}} \oplus \mathbf{U}_i \oplus \mathbf{H}_i)\right),
\end{equation}
where $\oplus$ is the concatenation operator.

The merits of the self-supervised subgraph identification method are twofold: 
\begin{enumerate}
    \item Different from most existing works using fixed hop or fixed size subgraphs which can not effectively capture long-range dependencies, our method fuses multi-view subgraphs to capture potential long-range dependencies.
    \item We identify subgraphs by maximizing mutual information in a self-supervised manner, and according to mutual information, each node in the subgraph has different fusion weights which enable subgraphs to perceive more informative nodes to augment representations.
\end{enumerate}

\subsection{Learning Objective}

The inductive bias of subgraph-based embedding augmentation is that more correlated neighbors can better represent the structural information of interested nodes. 
To leverage inductive bias between the naive subgraph embedding $\widetilde{\mathbf{S}}_{i}^{\mathbf{H}}$ and naive embedding $\mathbf{H}$, the latent subgraph embedding $\widetilde{\mathbf{S}}_{i}^{\mathbf{U}}$and latent embedding $\mathbf{U}$, and naive embedding $\mathbf{H}$ and latent embedding $\mathbf{U}$ to circumvent the issue of limited label information settings, we calculate the prototypical loss by aligning three pairs of prototypes, \emph{i.e.}, (1) prototypical naive subgraph $\mathbf{P}_{\widetilde{\mathbf{S}}^{\mathbf{H}}}$ and prototypical naive node $\mathbf{P}_{\mathbf{H}}$, (2) prototypical latent subgraph $\mathbf{P}_{\widetilde{\mathbf{S}}^{\mathbf{U}}}$ and prototypical latent node $\mathbf{P}_{\mathbf{U}}$, and (3) prototypical naive node $\mathbf{P}_{\mathbf{H}}$ and prototypical latent node $\mathbf{P}_{\mathbf{U}}$. The above four prototypes are obtained by taking the mean over all embedding of the same class,
\begin{equation}\label{eq:prototype}
\begin{split}
    \mathbf{P}^{(k)}_{\widetilde{\mathbf{S}}^{\Psi}}=\frac{1}{|\mathbf{V}_l^{(k)}|}\sum_{i\in \mathbf{V}_l^{(k)} } \widetilde{\mathbf{S}}_i^{\Psi},\\
    \mathbf{P}^{(k)}_{\Psi}=\frac{1}{|\mathbf{V}_l^{(k)}|}\sum_{i\in \mathbf{V}_l^{(k)} } \Psi_i,
\end{split}
\end{equation}
where $\mathbf{V}_l^{(k)}$ denotes the training set $\mathbf{V}_l$ of nodes belonging to class $k$.
The four prototypes serve as landmarks with respect to the inductive bias of itself. 
The prototypical loss can be calculated via the Euclidean distance between pairs of prototypes,

\begin{equation}\label{eq:al}
\mathcal{L}_{p}=\sum_{k=1}^K\left(
||\mathbf{P}^{(k)}_{\widetilde{\mathbf{S}}^{\mathbf{H}}}-\mathbf{P}^{(k)}_{\mathbf{H}}||
+||\mathbf{P}^{(k)}_{\widetilde{\mathbf{S}}^{\mathbf{U}}}-\mathbf{P}^{(k)}_{\mathbf{U}}||
+||\mathbf{P}^{(k)}_{\mathbf{H}}-\mathbf{P}^{(k)}_{\mathbf{U}}||\right).
\end{equation}

The augmented embedding $\overline{\mathbf{H}}_i$ is used for node classification. 
For predicting the probability of each class for each node, a softmax activation function is further performed on the embedding $\overline{\mathbf{H}}_i$, \emph{i.e.}, $\overline{\mathbf{H}}_i=softmax(\overline{\mathbf{H}}_i)$. 
The cross-entropy loss of predicted probability over all the labeled nodes $\mathbf{V}_l$ is minimized as follows,
\begin{equation}\label{eq:dynamics}
\mathcal{L}_{c}=-\frac{1}{|\mathbf{V}_l|}\sum_{i\in \mathbf{V}_l}\sum_{k=1}^{K}\mathbb{I}(y_i=k)\log(\overline{\mathbf{H}}_{i}),
\end{equation}
where $\mathbb{I}(\cdot)$ is an indicator function (if $y_i$ is the class $k$, then $\mathbb{I}(y_i=k)=1$, otherwise $\mathbb{I}(y_i=k)=0$.

The total loss function is the weighted sum of the classification loss $\mathcal{L}_{c}$, and the prototypical loss $\mathcal{L}_{p}$,
\begin{equation}\label{eq:dynamics}
\mathcal{L}=\mathcal{L}_{c}+\lambda_{p}\mathcal{L}_{p},
\end{equation}
where $\lambda_{p}$ is used for controlling the degree of the prototypical loss. Our proposed algorithm is sketched in Algorithm 1.

\begin{algorithm} 
  \caption{{\em Muse}}
  \label{alg:Meta}
  \KwIn{$\mathcal{G}=(\mathbf{V},\mathbf{A},\mathbf{X})$}
  \KwOut{Classification results}
  Construct the latent graph $\mathcal{G}^{'}=(\mathbf{V},\mathbf{A}^{'},\mathbf{X})$ by Isomap\;
 \While{the maximum number of iterations is not reached}{
 Obtain $\mathbf{H}$ and $\mathbf{U}$ by layer-wise propagation in Eq. (\ref{eq:gcn1}) \;
 \For{each node $i\in\mathbf{V}_l$}{
  Optimize $\mathbf{M}_i^{\mathbf{H}}$ and $\mathbf{M}_i^{\mathbf{U}}$ with $s$ steps by Eq. (\ref{eq:finalentropy})\;
  Get subgraph enbedding $\widetilde{\mathbf{S}}_{i}^{\mathbf{H}}$ and $\widetilde{\mathbf{S}}_{i}^{\mathbf{U}}$ by Eq. (\ref{eq:subgraphs})\;
  Get fusion representation $\overline{\mathbf{H}}_i=\sigma\left(FC(\widetilde{\mathbf{S}}_{i}^{\mathbf{H}} \oplus \widetilde{\mathbf{S}}_{i}^{\mathbf{U}} \oplus \mathbf{U}_i \oplus \mathbf{H}_i)\right) $\;
 }
 Get prototypes according to Eq. (\ref{eq:prototype})
 $\mathcal{L}_{p}=\sum_{k=1}^K(
||\mathbf{P}^{(k)}_{\widetilde{\mathbf{S}}^{\mathbf{H}}}-\mathbf{P}^{(k)}_{\mathbf{H}}||
+||\mathbf{P}^{(k)}_{\widetilde{\mathbf{S}}^{\mathbf{U}}}-\mathbf{P}^{(k)}_{\mathbf{U}}||
+||\mathbf{P}^{(k)}_{\mathbf{H}}-\mathbf{P}^{(k)}_{\mathbf{U}}||)$\;
 $\mathcal{L}_{c}=-\frac{1}{|\mathbf{V}_l|}\sum_{i\in \mathbf{V}_l}\sum_{k=1}^{K}\mathbb{I}(y_i=k)\log(\overline{\mathbf{H}}_{i})$\;
 Update all parameters according to $\mathcal{L}=\mathcal{L}_{c}+\lambda_{p}\mathcal{L}_{p}$ \;
 }
Predict the labels of unlabeled nodes based on the trained
model\;
   \Return the prediction results 
\end{algorithm}

\subsection{Theoretical Analysis}  
In this section, we provide a theoretical analysis regarding the generalized error of the proposed {\em Muse} to illustrate the effectiveness of capturing complementary information from multi-view subgraph embedding. Firstly, we give some definitions to guide the proof.

\begin{definition}\label{Rademacher}
(Rademacher complexity\cite{bartlett2002rademacher}). Let $\mathcal{F}$ be a real-valued function class and $\{x_i\}_{i=1}^{N}$ be a set of random variables from a distribution $\mathcal{P}_x$ of a domain $\mathcal{X}$. Denote $\{\sigma_i\}_{i=1}^{N}$ be a set of independent Rademacher random variables with zero mean and unit standard deviation. The Rademacher complexity of $\mathcal{F}$ with respect to $\{x_i\}_{i=1}^{N}$ is defined as
\begin{equation}
\mathfrak{R}(\mathcal{F})=\mathbb{E}_{\sigma}\left[ \sup_{f\in\mathcal{F}}\frac{1}{N}\sum_{i=1}^{N}\sigma_i f(x_i)\right].
\end{equation}
\end{definition}
Rademacher complexity
measures the richness of the real-valued function class $\mathcal{F}$ w.r.t the probability distribution $\mathcal{P}_x$.

\begin{myThe}\label{rf}
(Rademacher complexity bound of neural networks\cite{lu2021rademacher}).
Assuming that the neural network has $d$ layers with parameter matrices $\mathbf{W}_1,...,\mathbf{W}_d$ that are at most $\mathbf{M}_1,...,\mathbf{M}_d$, and the activation functions are 1-Lipschitz, positive-homogeneous. Let $x$ is upper bounded by $B$, \emph{i.e.}, for any $x$, $||x||\leq B$, then, 
\begin{equation}
\mathfrak{R}(\mathcal{F})\leq\frac{B(\sqrt{2d\log 2} +1)\prod_{i=1}^{d} \mathbf{M}_i}{\sqrt{N}}.
\end{equation}
\end{myThe}

\begin{definition}\label{McDiamid}
(Extended McDiamid's Inequality\cite{mcdiarmid1989method}). Given independent domains $\mathcal{X}^{(k)} (1\leq k \leq K)$, for any $k$, let $\{x\}^{m_k}$ be $m_k$ independent random variables taking values from the domain $\mathcal{X}^{(k)}$. Assume that the function $H: \mathcal{X}^{(1)}\times ...\times \mathcal{X}^{(K)} \rightarrow \mathbb{R}$ satisfies the condition of bounded difference: for all $1\leq k \leq K$ and $1\leq i\leq m_k$,
\begin{equation}
\sup_{\{x\}^{m_1},...,\{x\}^{m_K},x_i\in \{x\}^{m_k}} |H-H^{'}|\leq c_i^{(k)},
\end{equation}
where $H=H(\{x\}^{m_1},...,\{x\}^{m_k},...,\{x\}^{m_K})$ and 
\begin{equation}
\begin{split}
 H^{'}=&H(\{x\}^{m_1},...,\{x\}^{m_{k-1}},...\\
&...,\{x_1,...,x_i^{'},...,x_{m_{k}}\}^{m_k},\{x\}^{m_{k+1}},...,\{x\}^{m_K}),\\
\end{split}
\end{equation}
Then, for any $\xi>0$, 
\begin{equation}
     \Pr \left( H-\mathbb{E}\left(H\right)\ge \xi \right)
     \leq \exp\left(-2\xi^2/\sum_{k=1}^{K}\sum_{i=1}^{m_k}(c_i^{(k)})^2\right).
\end{equation}
\end{definition}

Before formally proceeding, we define some notations for convenience. Given an input node that has $V$ views of can be provided, we define $V$ mapping function w.r.t the $V$ views as $h=(h_1,...,h_V)$. Specifically, in our proposed {\em Muse}, $V$ is equal to 2, because we have the function with two views $h=(h_1,h_2)$ to handle the data $x=(x^1,x^{2})\in \{x_i\}_{i=1}^{N}$ with respect to the views of the raw graph and the latent graph. The generalized error $\mathcal{R}$ of the {\em Muse} can be denoted as,
\begin{equation}
\mathcal{R}(h,x)=\frac{1}{V}\sum_{v=1}^{V}\frac{1}{N}\sum_{i=1}^{N}[\mathcal{L}(h_v(x_i^v),y_i))].
\end{equation}
For simplicity, we denote $\frac{1}{N}\sum_{i=1}^{N}[\mathcal{L}(h_v(x_i^v),y_i))]$ as $f_v(x^v)$, and $\mathcal{R}(h,x)$ can be rewritten as,
\begin{equation}
\mathcal{R}(f,x)=\frac{1}{V}\sum_{v=1}^{V}f_v(x^v).
\end{equation}

In the following part, we bound the generalization error by using Rademacher Complexity.

\begin{myThe}\label{proposed}
(Generalization bound of the proposed {\em Muse}). Assume that the function class $\mathcal{F}$ is bounded by $[a,b]$, and parameters $\mathbf{W}_1,...,\mathbf{W}_d$ of the proposed neural network are at most $\mathbf{M}_1,...,\mathbf{M}_d$, and the activation functions are 1-Lipschitz, positive-homogeneous. Let $x=(x^1,...,x^V)$ has $V$ views of representations and $ x\in \{x_i\}_{i=1}^{N} \sim \mathcal{X}$ is upper bounded by $B$, \emph{i.e.}, for any $x$, $||x||\leq B$. For any $\delta\in (0,1)$, then with probability at least $1-\delta$ over $\mathcal{X}$, there holds that for any $f\in \mathcal{F}$,
\begin{equation}
\begin{split}
\mathbb{E}_{x\sim\mathcal{X}}[{\mathcal{R}(f,x)}]\leq \mathcal{R}(f,x)  + & \frac{2B(\sqrt{2d\log 2} +1)\prod_{i=1}^{d} \mathbf{M}_i}{\sqrt{N}}\\
 + &\sqrt{\frac{(b-a)^2\ln(4/\delta)}{2VN}}.
 \end{split}
\end{equation}
\end{myThe}

Next, we begin to prove Theorem \ref{proposed} by defining the following equation according to the extended McDiamid's inequality,
\begin{equation}\label{hprno}
H(\mathcal{X}^{(1)},...,\mathcal{X}^{(V)})=\sup_{f\in\mathcal{F}}\left[\mathbb{E}_{x\sim \mathcal{X}} R(f,x)-R(f,x)\right],
\end{equation}
Specifically, for the proposed {\em Muse}, $x=(x^1,x^2)\in\mathcal{X}$ has two views of representations with respect to the raw graph domain $\mathcal{X}^{(1)}$ and the latent graph domain $\mathcal{X}^{(2)}$, respectively. $H(\mathcal{X}^{(1)},...,\mathcal{X}^{(V)})$ satisfies the condition of bounded difference with
\begin{equation}
c_i^{(1)}=c_i^{(2)}=...=c_i^{(V)}=\frac{(b-a)}{VN}.
\end{equation}
Equivalently, with probability at least $1-(\delta/4)$,
\begin{equation}\label{tail}
\begin{split}
    H(\mathcal{X}^{(1)},...,\mathcal{X}^{(V)}) & \leq  \mathbb{E}_{x}\left(H(\mathcal{X}^{(1)},...,\mathcal{X}^{(V)})\right)  \\ &+\sqrt{\frac{(b-a)^2\ln(4/\delta)}{2VN}}.
    \end{split}
\end{equation}
Based on Jensen’s inequality, and Definition \ref{Rademacher}, for any two  nodes $x$ and $x^{'}$, we have
\begin{equation}
\begin{split}
    \mathbb{E}_{x}\left(H(\mathcal{X}^{(1)},...,\mathcal{X}^{(V)})\right) & =\mathbb{E}_{x}\left(\sup_{f\in\mathcal{F}}\mathbb{E}_{x} R(f,x)-R(f,x)\right)\\
 \leq & \mathbb{E}_{x}\left(\sup \frac{1}{V}\sum_{v=1}^{V}f_v(x^v)-f_v(x^{'v})\right).
\end{split}
\end{equation}

Given a set of independent variables $\{\sigma_v\}_{v=1}^{V}$,  uniformly distributed on $\{-1,1\}$, we define
\begin{equation}
g_{\sigma_v}(x,x^{'})=
\begin{cases}
x \ \ \mathbf{i}\mathbf{f} \  \ \sigma_v=1,\\ 
x^{'} \ \ \mathbf{i}\mathbf{f} \  \ \sigma_v=-1,
\end{cases}
\end{equation}
and 
\begin{equation}
g_{\sigma_v}^{’}(x^{'},x)=
\begin{cases}
x \ \ \mathbf{i}\mathbf{f} \  \ \sigma_v=-1,\\ 
x^{'} \ \ \mathbf{i}\mathbf{f} \  \ \sigma_v=1,
\end{cases}
\end{equation}
Then we can have  
\begin{equation}\label{2r}
\begin{split}
& \mathbb{E}_x\left(\sup \frac{1}{V}\sum_{v=1}^{V}f_v(x^v)-f_v(x^{'v})\right) \\
 = &\mathbb{E}_{\sigma}\left[\mathbb{E}_{x}\left[\sup_{f\in\mathcal{F}}\frac{1}{V}\sum_{v=1}^{V}f_v\left(g^{'}(x^{v},x^{'v})\right)-f_v\left(g(x^{v},x^{'v})\right)|\sigma\right]\right]\\
 =&\mathbb{E}_{\sigma,x}\left[\sup_{f\in\mathcal{F}}   \sum_{v=1}^{V}\sigma_v\left(f_v(x^{'v})-f_v(x^{v})\right)\right]\\
 \leq& 2\mathbb{E}_{\sigma,x}\left(\sup_{f\in\mathcal{F}}   \sum_{v=1}^{V}\sigma_v f_v(x^{'v})\right)=2 \mathfrak{R}(\mathcal{F}),
\end{split}
\end{equation}
By combining Eq. (\ref{tail}), Eq. (\ref{2r}) and Theorem \ref{rf}, we can get
\begin{equation}
\sup_{f\in\mathcal{F}}\left[\mathbb{E}_{x\sim \mathcal{X}} R(f,x)-R(f,x)\right]\leq 2 \mathfrak{R}(\mathcal{F})+\sqrt{\frac{(b-a)^2\ln(4/\delta)}{2VN}}.
\end{equation}
Therefore, we complete the proof.

Theorem \ref{proposed} indicates that the generalization error is bounded by the empirical training risk, Rademacher complexity, and the additional error. The empirical training risk and Rademacher complexity are caused by the finite samples. As the sample size tends to infinity, the empirical training risk and Rademacher complexity tend to be zero. From the second term, the generalization error bound indicates that when the number of distinct views of data $V$ increases, the additional error can be further reduced. In our work, we provide two views of subgraph-level representations, thus improving the performance in handling graph data with scarce labeled nodes.

\section{Computational Complexity}

In this section, we give the time complexity of \emph{Muse}. Generating the latent embedding and naive embedding requires the cost of $\mathcal{O}(\sum_{l=1}^{L} n_{l}n_{l+1}|\mathcal{E}|)$, where $L$ is the number of layers in the GNN, $n_{l}$ is the number of neurons of $l$-th layer and $|\mathcal{E}|$ is the number of edges. To generate the latent subgraph embedding and naive subgraph embedding,  we use gradient descent to optimize the mask vector of each node, which requires an additional cost of $\mathcal{O}(|\mathbf{V}|^2)$.

\section{Experimental Studies}

In this section, we will start with the experimental settings and then show our experiment results to answer the following questions:\\ 
\textbf{Q1}: {\em Can the proposed {\em Muse} achieve promising results in solving the node classification tasks with scarce labeled data?}\\
\textbf{Q2}: {\em Can {\em Muse} capture the complementary information by extracting the latent subgraph embedding and the naive subgraph embedding?}\\
\textbf{Q3}: {\em How does each component in the proposed {\em Muse} affect the performance?}\\
\textbf{Q4}: {\em How do hyper-parameters affect performance?}



\subsection{Benchmark Datasets}

\zz{All methods are evaluated on five datasets, including three citation networks and two social networks}. Cora~\cite{sen2008collective}, Citeseer~\cite{sen2008collective}, Pubmed~\cite{sen2008collective} are citation networks, where each node is a document and the edges describe the citation relationships. A document is assigned a unique label based on its topic. Node features are bag-of-words representations of the documents. 
BlogCatalog~\cite{meng2019co} is a social network, where node features are generated by users as a short description of their blogs. The labels denote the topic categories provided by the authors.
\zz{Flickr~\cite{huang2017label} is an image-sharing-based social network, where each node is a user, edges are the interaction records between users, and the users are labeled with the joined groups.} The statistical details of the four graph datasets are listed in Table \ref{tab:datasets}.

\begin{table}[htbp]
  \centering
  \caption{Detailed information of the four graph datasets, where ``\#" denotes ``the number of".}
  \resizebox{89mm}{10mm}{
  \textcolor{black}{
    \begin{tabular}{cccccc}
    \toprule
    Property & Cora  & Citeseer & Pubmed & BlogCatalog & Flickr \\
    \midrule
    \# Nodes & 2,708 & 3,327 & 19,717 & 5,196 & 7,575 \\
    \# Edges & 10,556 & 9,228 & 44,338 & 171,743 & 487,051 \\
    \# Features & 1,433 & 3,703 & 500 & 8,189 & 12,047 \\
    \# Labels & 7     & 6     & 3 & 6 & 9\\
    \bottomrule
    \end{tabular}%
    }
    }
  \label{tab:datasets}%
\end{table}%

\begin{table*}[htbp]
\small
  \centering
  \caption{Classification accuracy on three citation networks and one social network (\%).}
   \resizebox{180mm}{27mm}{
\textcolor{black}{
    \begin{tabular}{ccccc|cccc|cccc}
    \toprule
    \multirow{2}[3]{*}{Algorithm} & \multicolumn{4}{c|}{1 label per class} & \multicolumn{4}{c|}{2 labels per class} & \multicolumn{3}{c}{5 labels per class} &  \\
\cmidrule{2-13}          & Cora  & Citeseer & Pubmed & BlogCatalog & Cora  & Citeseer & Pubmed & BlogCatalog & Cora  & Citeseer & Pubmed & BlogCatalog \\
\midrule
    MLP   & 41.4$\pm$0.4 & 30.5$\pm$0.9 & 47.0$\pm$1.0 & 37.0$\pm$0.5 & 48.4$\pm$0.7 & 33.6$\pm$0.8 & 50.8$\pm$0.5 & 43.7$\pm$0.6 & 55.0$\pm$0.8 & 40.2$\pm$0.8 & 59.8$\pm$0.9 & 52.2$\pm$0.8 \\
    GCN   & 45.5$\pm$0.8 & 31.2$\pm$1.4 & 49.2$\pm$1.5 & 43.9$\pm$1.3 & 53.9$\pm$0.8 & 38.7$\pm$0.7 & 55.4$\pm$1.0 & 46.6$\pm$1.0 & 67.0$\pm$0.9 & 54.9$\pm$1.2 & 67.8$\pm$3.0 & 55.3$\pm$1.6 \\
    GAT   & 46.3$\pm$0.5 & 32.0$\pm$1.2 & 51.0$\pm$1.6 & 23.5$\pm$1.4 & 55.0$\pm$1.3 & 40.4$\pm$0.4 & 57.2$\pm$1.8 & 24.7$\pm$1.4 & 69.0$\pm$0.6 & 55.9$\pm$0.9 & 66.5$\pm$1.5 & 36.5$\pm$1.7 \\
    GraphSAGE & 45.0$\pm$0.6 & 32.3$\pm$0.9 & 52.6$\pm$2.2 & 42.1$\pm$1.6 & 54.1$\pm$1.5 & 39.0$\pm$0.9 & 57.6$\pm$1.8 & 44.9$\pm$1.3 & 65.5$\pm$0.8 & 55.0$\pm$0.8 & 67.0$\pm$2.0 & 52.2$\pm$1.2 \\
    SGC   & 44.2$\pm$0.6 & 31.8$\pm$0.8 & 53.4$\pm$1.4 & 44.3$\pm$1.2 & 53.6$\pm$0.8 & 39.8$\pm$1.0 & 56.5$\pm$2.0 & 47.4$\pm$1.0 & 66.0$\pm$1.0 & 54.6$\pm$1.0 & 66.4$\pm$2.3 & 55.5$\pm$1.6 \\
    DAGNN & 60.2$\pm$1.5 & 50.2$\pm$0.9 & 60.5$\pm$2.2 & 41.3$\pm$1.0 & 67.0$\pm$1.0 & 57.0$\pm$1.6 & 66.9$\pm$2.0 & 45.4$\pm$1.1 & 71.0$\pm$1.4 & 59.0$\pm$0.9 & 68.9$\pm$1.4 & 60.7$\pm$1.3 \\
    APPNP & 55.8$\pm$0.7 & 43.9$\pm$1.8 & 53.0$\pm$2.1 & 44.5$\pm$1.8 & 58.9$\pm$0.8 & 48.4$\pm$0.9 & 57.7$\pm$1.5 & 47.3$\pm$1.5 & 62.5$\pm$0.9 & 57.2$\pm$1.2 & 64.8$\pm$0.9 & 60.3$\pm$1.6 \\
    ICGN  & 43.6$\pm$0.9 & 40.5$\pm$1.6 & 52.0$\pm$1.0 & 40.8$\pm$2.0 & 53.0$\pm$1.7 & 44.0$\pm$1.5 & 58.5$\pm$2.5 & 41.0$\pm$2.5 & 62.6$\pm$2.1 & 58.0$\pm$1.5 & 65.0$\pm$1.0 & 45.0$\pm$2.2 \\
    Shoestring & 61.6$\pm$1.3 & 55.8$\pm$1.4 & 61.5$\pm$3.0 & 44.7$\pm$1.9 & 67.0$\pm$1.8 & 62.5$\pm$1.0 & 66.0$\pm$2.2 & 45.9$\pm$1.3 & 70.8$\pm$1.0 & 64.8$\pm$1.8 & 66.5$\pm$2.3 & 45.8$\pm$1.5 \\
        GraphHop & 44.9$\pm$1.7 & 37.7$\pm$1.4 & 49.6$\pm$1.9 & 40.4$\pm$1.4 & 54.5$\pm$1.7 & 48.9$\pm$2.2 & 60.0$\pm$2.0 & 44.3$\pm$1.2 & 59.0$\pm$1.5 & 49.0$\pm$1.4 & 60.9$\pm$2.2 & 56.0$\pm$1.7 \\
            \midrule
    SUBG-CON & 59.8$\pm$1.0 & 37.6$\pm$1.5 & 52.8$\pm$5.0 & 43.6$\pm$2.7 & 63.4$\pm$2.6 & 40.3$\pm$3.1 & 59.4$\pm$4.6 & 46.7$\pm$3.5 & 68.4$\pm$3.0 & 49.9$\pm$3.5 & 64.1$\pm$5.1 & 55.8$\pm$3.8 \\
    SelfSAGCN & 65.2$\pm$0.9 & 50.4$\pm$3.3 & 65.3$\pm$3.0 & 45.8$\pm$3.9 & 72.0$\pm$0.8 & 66.5$\pm$1.4 & 65.9$\pm$2.0 & 47.5$\pm$3.0 & \boldmath{}\textbf{77.9$\pm$0.8}\unboldmath{} & 65.5$\pm$1.5 & 70.8$\pm$1.6 & 56.7$\pm$2.7 \\
    SCRL  & 60.8$\pm$1.3 & 58.0$\pm$2.0 & 66.1$\pm$3.8 & 40.5$\pm$2.3 & 68.1$\pm$2.0 & 67.5$\pm$2.5 & 68.2$\pm$3.6 & 45.3$\pm$2.5 & 65.3$\pm$2.8 & 71.6$\pm$3.4 & 70.4$\pm$3.3 & 53.5$\pm$3.5 \\
    NAGphormer & 52.6$\pm$1.4 & 36.0$\pm$1.6 & 65.6$\pm$2.5 & 44.4$\pm$2.3 & 62.5$\pm$0.9 & 47.7$\pm$1.8 & 68.1$\pm$2.3 & \boldmath{}\textbf{52.5$\pm$1.3}\unboldmath{} & 73.5$\pm$0.9 & 52.5$\pm$1.5 & 71.3$\pm$1.9 & \boldmath{}\textbf{65.0$\pm$1.3}\unboldmath{} \\
    \midrule
    {\em Muse} & \boldmath{}\textbf{69.5$\pm$0.7}\unboldmath{} & \boldmath{}\textbf{59.6$\pm$1.6}\unboldmath{} & \boldmath{}\textbf{67.4$\pm$2.4}\unboldmath{} & \boldmath{}\textbf{48.7$\pm$2.0}\unboldmath{} & \boldmath{}\textbf{73.8$\pm$1.7}\unboldmath{} & \boldmath{}\textbf{69.5$\pm$2.0}\unboldmath{} & \boldmath{}\textbf{69.6$\pm$2.4}\unboldmath{} & 49.6$\pm$2.2 & 77.4$\pm$2.3 & \boldmath{}\textbf{72.3$\pm$2.5}\unboldmath{} & \boldmath{}\textbf{72.6$\pm$2.8}\unboldmath{} & 62.0$\pm$1.7 \\
    \bottomrule
     \end{tabular}%
     }
  \label{tab:compare}%
  }
\end{table*}%

\begin{table}[htbp]
  \centering
  \caption{Classification accuracy on the Flickr dataset under different numbers of labeled nodes. (\%).}
   \resizebox{90mm}{26.0mm}{
      \textcolor{black}{
    \begin{tabular}{ccccc}
    \toprule
    \multirow{2}[3]{*}{Algorithm} & \multicolumn{4}{c}{Flickr} \\
\cmidrule{2-5}          & 1 label per class & 2 labels per class & 5 labels per class & 20 labels per class \\
    \midrule
    MLP   & 11.3$\pm$0.5 & 11.5$\pm$0.9 & 11.6$\pm$0.8 & 12.6$\pm$0.7 \\
    GCN   & 12.4$\pm$0.7 & 14.5$\pm$0.7 & 15.4$\pm$0.5 & 17.2$\pm$0.5 \\
    GAT   & 13.6$\pm$0.3 & 14.4$\pm$0.4 & 15.5$\pm$0.5 & 17.5$\pm$0.8 \\
    GraphSAGE & 12.6$\pm$0.8 & 14.5$\pm$0.9 & 15.7$\pm$0.7 & 17.0$\pm$0.9 \\
    SGC   & 12.1$\pm$0.8 & 13.8$\pm$0.6 & 15.3$\pm$0.4 & 17.1$\pm$0.6 \\
    DAGNN & 15.0$\pm$0.4 & 16.2$\pm$0.5 & 17.0$\pm$0.6 & 18.2$\pm$0.9 \\
    APPNP & 11.4$\pm$0.9 & 11.5$\pm$0.8 & 11.9$\pm$0.6 & 13.8$\pm$0.8 \\
    ICGN  & 14.1$\pm$0.5 & 11.4$\pm$0.4 & 15.1$\pm$0.7 & 15.9$\pm$0.5 \\
    Shoestring & 14.4$\pm$0.6 & 14.7$\pm$0.7 & 15.4$\pm$0.8 & 16.2$\pm$0.5 \\
    GraphHop & 11.7$\pm$0.3 & 11.7$\pm$0.3 & 12.0$\pm$0.4 & 15.1$\pm$0.8 \\
    \midrule
    SUBG-CON & 13.0$\pm$0.5 & 13.4$\pm$0.4 & 15.2$\pm$0.5 & 16.2$\pm$0.7 \\
    SelfSAGCN & 17.3$\pm$1.1 & 21.2$\pm$0.8 & 25.1$\pm$0.9 & 28.0$\pm$0.7 \\
    SCRL  & 12.4$\pm$0.6 & 13.6$\pm$0.6 & 13.8$\pm$0.8 & 15.5$\pm$0.7 \\
    NAGphormer & 20.4$\pm$1.0 & 22.9$\pm$0.9 & 32.0$\pm$1.1 & \boldmath{}\textbf{43.3$\pm$1.3}\unboldmath{} \\
    \midrule
    {\em Muse} & \boldmath{}\textbf{22.5$\pm$0.5}\unboldmath{} & \boldmath{}\textbf{23.5$\pm$0.5}\unboldmath{} & \boldmath{}\textbf{33.7$\pm$0.4}\unboldmath{} & 37.0$\pm$0.6 \\
    \bottomrule
    \end{tabular}%
    }
    }
  \label{tab:flickr}%
\end{table}%

\subsection{Baselines}

For performance comparison of node classification with scarce labeled nodes, in addition to Multilayer Perceptron (MLP), two categories of baselines are considered: semi-supervised learning methods and self-supervised learning methods.

(1) Semi-supervised learning methods: We included the following state of arts with scarce label learning:
    DAGNN~\cite{DAGNN} attempts to enlarge receptive fields to overcome over-smoothing when the number of training nodes is limited under the semi-supervised learning setting.
    APPNP~\cite{APPNP} incorporates GCN with personalized PageRank to achieve efficient propagation for semi-supervised classification.
    ICGN~\cite{ICGN} uses flexible graph filtering for efficient label learning with few labels.
    Shoestring~\cite{9157520} incorporates metric learning networks to graph-based semi-supervised learning for severely limited labeled nodes.
    \zz{GraphHop~\cite{9737682} is a smoothening label propagation algorithm, in which each propagation alternates between label aggregation and label update.}

Moreover, for a comprehensive understanding of the effectiveness of our method, we also consider several classic graph-based semi-supervised learning methods. They are Graph Convolutional Network (GCN)~\cite{10GraphConvolutionalNetworks}, Graph Attention Network (GAT)~\cite{GATvelivckovic2017graph},GraphSAGE~\cite{10GraphSAGE}, and SGC~\cite{wu2019simplifying}.

(2) Self-supervised learning methods: 
    SUBG-CON~\cite{jiao2020sub} is a contrastive-based method, where positive and negative subgraphs are sampled to induce a contrastive loss. 
    
    In addition to the contrastive-based method, three context-based methods including SelfSAGCN~\cite{SelfSAGCNCVPR}, SCRL~\cite{liu2021self}, and NAGphormer~\cite{chen2023nagphormer} are selected for the comparison.
    SelfSAGCN seeks extra semantic information from nodes and serves the semantic features as supervised signals.
    SCRL integrates correlated information from both graph structure and node features to maintain the consistency of features and topology. 
         \zz{NAGphormer~\cite{chen2023nagphormer} transforms features of neighborhoods into tokens and treats each node as a sequence of tokens to preserve graph structural information.}

\subsection{Experimental Setup}

We randomly select nodes for each class from the training set for training the models. In our experimental setup, the number of trials is set to 10, and the mean results over 10 random trials of reproduced baselines are reported. More specifically, for each dataset, we test the comparison algorithms on four scenarios, \emph{i.e.}, 1 labeled node per class, 2 labeled nodes per class, 5 labeled nodes per class, and 20 labeled nodes per class, to cover the test cases of limited data and sufficient data.


Parameter settings are as follows: Two layers of GNN are built, and the number of hidden units is 17. ReLU is adopted as the non-linear activation, and softmax is used as the last layer for classification.
We employ the Adam SGD optimizer with a learning rate of 0.01, 0.5 dropout rate, $5 \times 10^{-4}$ weight decay. $\lambda_p$ is set as $4$.
The threshold values $k$ and $\tau$ for masking are set to 3 and 0.5, respectively, and the number of steps $s$ for optimizing masks is 20. 

\subsection{Results and Discussions}

To examine the performance of {\em Muse} (\textbf{Q1}), we evaluate the proposed {\em Muse} and all the baseline models on different node classification tasks with various number of labeled nodes. The experimental results on Cora, Citeseer, Pubmed, and BlogCatalog datasets are presented in Table \ref{tab:compare}, \zz{and the results on Flickr datasets are shown in Table \ref{tab:flickr}}. We highlighted the top classification accuracy in bold. The experimental findings are summarized as follows.

\begin{figure*}[htbp] 
  \centering   
  \subcaptionbox{GCN}{
  \includegraphics[width=3.65cm,height=3.99cm]{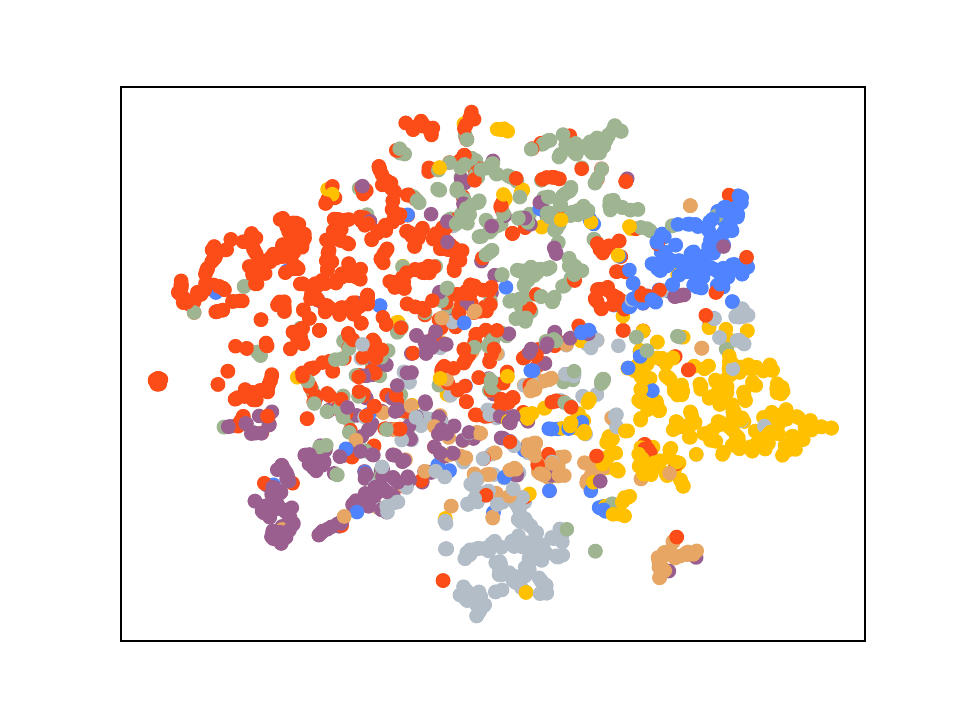}
  }
  \quad 
  \subcaptionbox{GCN-{\em Muse}}{
  \includegraphics[width=3.65cm,height=3.99cm]{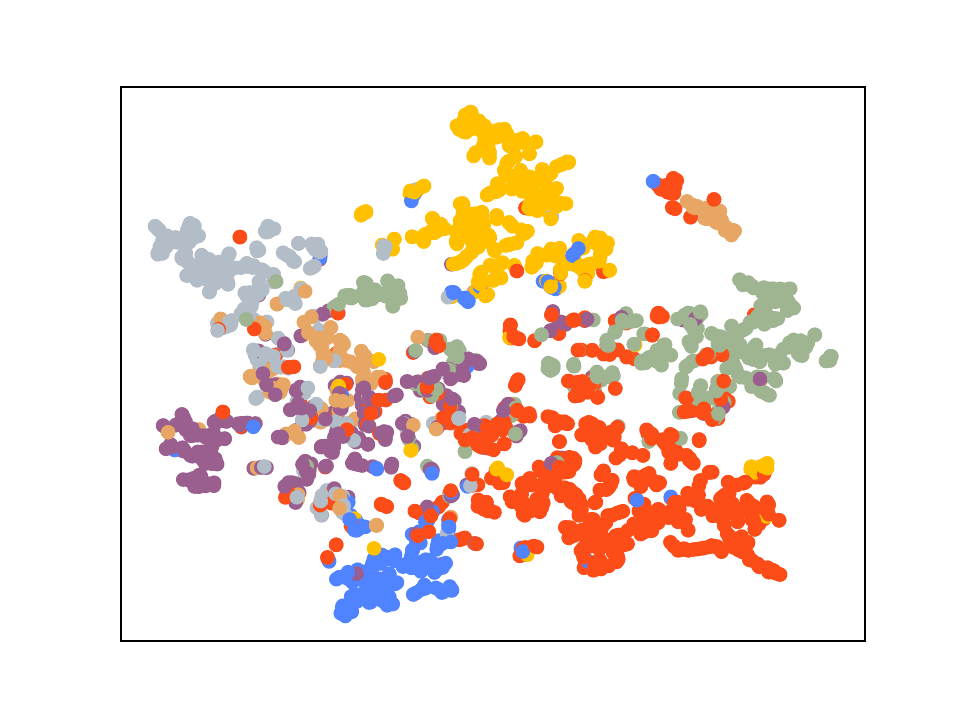}
  }
  \quad
  \subcaptionbox{GraphSAGE}{
  \includegraphics[width=3.65cm,height=3.99cm]{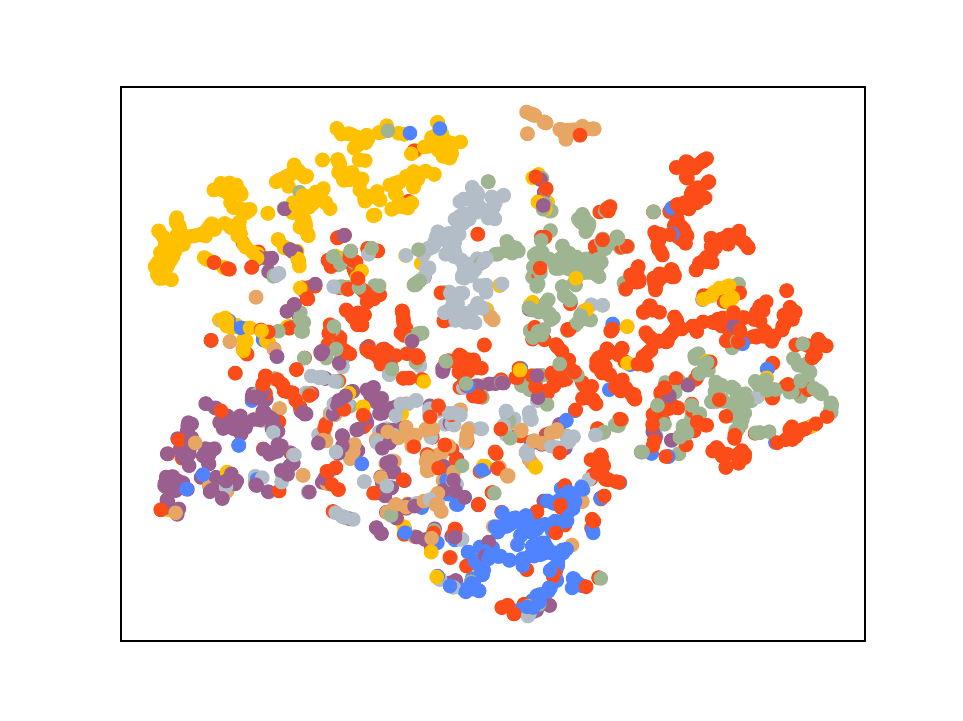} 
  }
  \quad
  \subcaptionbox{GraphSAGE-{\em Muse}}{
  \includegraphics[width=3.65cm,height=3.99cm]{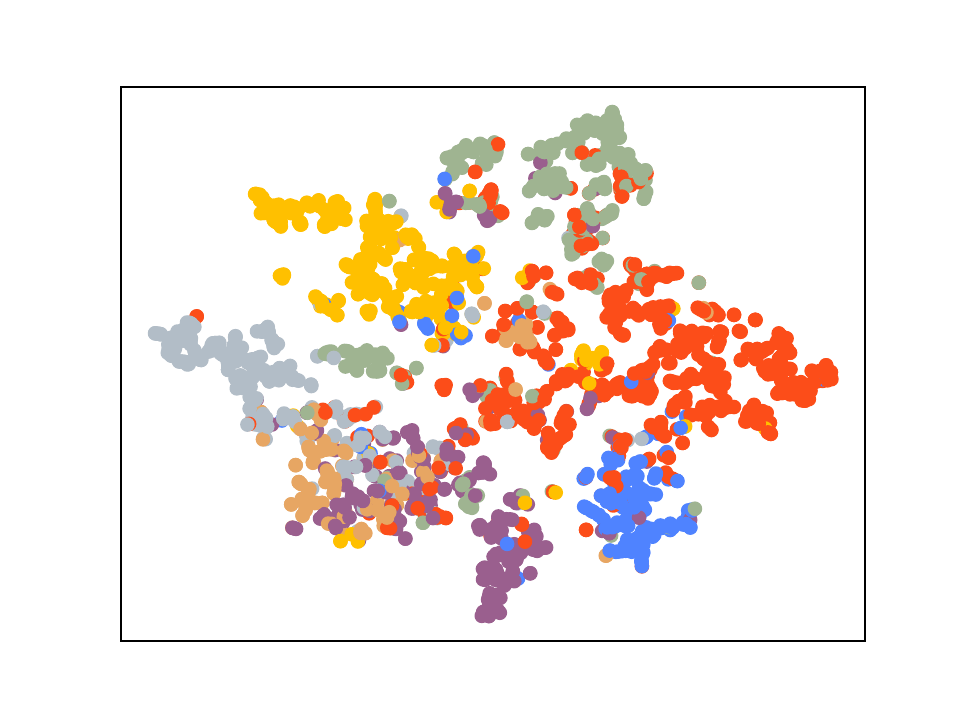}
  }
  \caption{t-SNE visualization of features derived by GCN, GCN-{\em Muse}, GraphSAGE, and GraphSAGE-{\em Muse} under 1 label per class setting. The features learned by GCN-{\em Muse} and GraphSAGE-{\em Muse} have compact clusters and clear boundaries.}
  \label{fig:tsne}
\end{figure*} 

\begin{figure}[h]
  \centering
  \includegraphics[width=9.1cm]{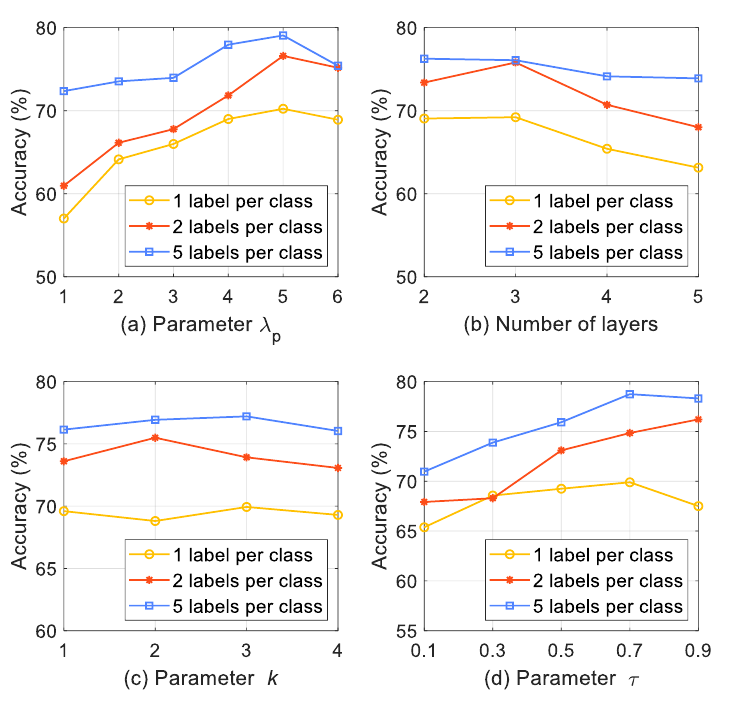}
  \caption{Document classification accuracy with different hyper-parameters on the Cora dataset. (a) Results with different parameters $\lambda_p$. (b) Results with different numbers of layers. (c) Results with different parameters $k$. (d) Results with different parameters $\tau$. }
  \label{fig:parameter}
\end{figure}

\zz{The experimental results from Table \ref{tab:compare} and Table \ref{tab:flickr} show that {\em Muse} achieves the best performance for 12 out of 16 cases}, indicating the advantage of fusing two views of subgraphs to augment the representations of nodes.
The reason for the high performance of the proposed {\em Muse} in handling the scenario of the severely limited labeled samples is that {\em Muse} can capture both useful information from the local context and global context based on the manifold assumption, which makes it can transfer as much knowledge as possible from limited labeled nodes to a large number of unlabeled nodes.

Secondly, SSL methods have shown great advantages over both supervised learning methods and semi-supervised learning methods.
Supervised learning methods face severe overfitting to the labeled nodes. Because the number of the labeled nodes working as “anchors” is too low, the information extracted from these labeled nodes cannot be effectively propagated to unlabeled samples, thus leading to poor generalization.
Because graph data is non-i.i.d, capturing inter-dependency among nodes can bring informative knowledge. Therefore, compared with semi-supervised learning methods, SSL methods can capture more information from unlabeled nodes by capturing inter-dependency.


Thirdly, we can observe that the proposed {\em Muse} gains more improvements on 1 label per class than with 5 labels per class. This is because when the labeled nodes are extremely scarce, the unlabeled nodes are more distant from the labeled nodes with homophily properties. In such a case, long-range dependencies enable homophily properties of labeled nodes can be captured as the supervisory signals, thus significantly improving the performance.

Lastly, all SSL and semi-supervised learning approaches consistently perform better than supervised learning methods. The difference implies again that traditional supervised learning methods are not designed to work on the scarce labeled node settings. From the experimental results shown in Table \ref{tab:compare}, although the number of labels is modestly increased, the improvement of traditional supervised learning methods is very limited.

\begin{figure}[!t]
  \centering
  \includegraphics[width=8.1cm]{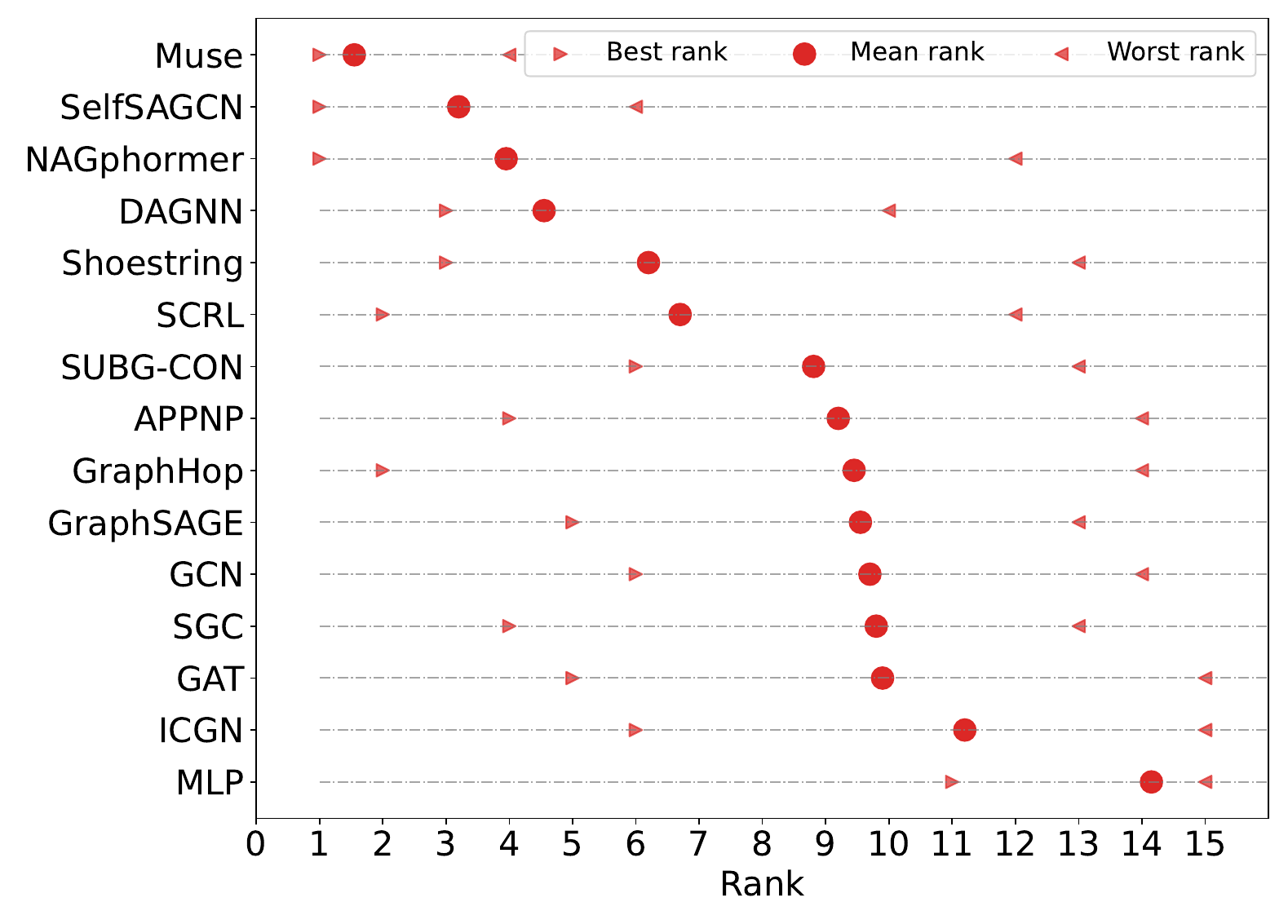}
  \caption{\zz{The best, worst, and average ranks of compared algorithms on the five datasets under all settings of numbers of labeled nodes.}}
  \label{fig:rank}
\end{figure}

\zz{We further conducted experiments to validate the performance under more labeled data (20 labels per class). The results in Table \ref{tab:20class} and Table \ref{tab:flickr} indicate {\em Muse} still outperforms most methods when the number of labeled nodes is sufficient. However, {\em Muse} falls slightly short compared to NAGphormer, a Transformer-based self-supervised model. Although proposed {\em Muse} is slightly behind NAGphormer under sufficient labeled samples, it is essential to emphasize that our algorithm primarily targets low-data scenarios and surpasses most methods in these scenarios. To provide a comprehensive overview of algorithm performance across the five datasets at varying numbers of labeled nodes, we conducted a statistical analysis to show the best, worst, and average rankings among these algorithms. As depicted in Fig. \ref{fig:rank}, our proposed {\em Muse} achieves the top ranking and the ranking is more stable compared to other algorithms.}

\begin{table}[htbp]
  \centering
  \caption{Classification accuracy on three citation networks and one social network with a sufficient number of labeled data(\%).}
   \resizebox{90mm}{30.0mm}{
   \textcolor{black}{
    \begin{tabular}{ccccc}
    \toprule
    \multirow{2}[3]{*}{Algorithm} & \multicolumn{4}{c}{20 labels per class} \\
\cmidrule{2-5}          & Cora  & Citeseer & Pubmed & BlogCatalog \\
    \midrule
    MLP   & 62.4$\pm$1.3 & 47.2$\pm$1.2 & 63.9$\pm$1.7 & 60.1$\pm$1.5 \\
    GCN   & 69.9$\pm$1.4 & 60.5$\pm$1.5 & 69.8$\pm$2.8 & 65.9$\pm$1.8 \\
    GAT   & 73.0$\pm$1.7 & 60.4$\pm$1.4 & 70.5$\pm$1.3 & 45.5$\pm$1.9 \\
    GraphSAGE & 72.6$\pm$1.1 & 61.0$\pm$0.9 & 69.7$\pm$1.6 & 62.0$\pm$1.3 \\
    SGC   & 70.2$\pm$1.5 & 59.7$\pm$0.9 & 70.7$\pm$2.0 & 63.1$\pm$1.5 \\
    DAGNN & 73.7$\pm$1.2 & 63.7$\pm$1.4 & 71.0$\pm$1.9 & 67.3$\pm$1.3 \\
    APPNP & 70.3$\pm$1.8 & 66.2$\pm$1.5 & 69.2$\pm$1.7 & 62.9$\pm$0.7 \\
    ICGN  & 66.5$\pm$2.0 & 62.0$\pm$1.6 & 69.0$\pm$1.3 & 47.9$\pm$2.1 \\
    Shoestring & 72.7$\pm$1.0 & 66.8$\pm$1.5 & 71.5$\pm$1.7 & 54.0$\pm$1.0 \\
        GraphHop & 79.6$\pm$0.6 & 68.0$\pm$1.0 & 76.4$\pm$1.5 & \boldmath{}\textbf{73.3$\pm$1.2}\unboldmath{} \\
    SelfSAGCN & 79.5$\pm$0.9 & 70.8$\pm$2.2 & 72.7$\pm$1.4 & 68.1$\pm$2.5 \\
    SCRL  & 70.8$\pm$1.2 & 73.0$\pm$2.2 & 72.1$\pm$2.9 & 60.5$\pm$1.9 \\
    NAGphormer & \boldmath{}\textbf{80.6$\pm$0.7}\unboldmath{} & 67.0$\pm$1.4 & \boldmath{}\textbf{76.7$\pm$1.6}\unboldmath{} & 73.1$\pm$1.5 \\
    \midrule
    {\em Muse} & 78.8$\pm$0.5 & \boldmath{}\textbf{73.5$\pm$1.4}\unboldmath{} & 73.6$\pm$3.1 & 72.4$\pm$1.5 \\
    \bottomrule
    \end{tabular}%
    }
    }
  \label{tab:20class}%
\end{table}%

\subsection{Ablation Study}

\renewcommand{\arraystretch}{0.75}
\begin{table*}[htbp]
  \centering
  \caption{Classification accuracy of various ablation algorithms on three citation networks and one social network (\%).}
     \resizebox{180mm}{44mm}{
    \begin{tabular}{ccccccccc}
    \toprule
    \multirow{2}[4]{*}{Algorithm} & \multicolumn{4}{c}{1 label per class} & \multicolumn{3}{c}{5 labels per class} &  \\
\cmidrule{2-9}          & Cora  & Citeseer & Pubmed & BlogCatalog & Cora  & Citeseer & Pubmed & BlogCatalog \\
    \midrule
    GCN ($\mathcal{G}$) & 43.1$\pm$0.7 & 29.2$\pm$0.9 & 51.5$\pm$1.8 & 43.9$\pm$1.3 & 69.1$\pm$0.5 & 53.7$\pm$0.9 & 68.4$\pm$3.2 & 55.3$\pm$1.6 \\
    GCN ($\mathcal{G}^{'}$) & 43.0$\pm$1.2 & 28.0$\pm$0.8 & 52.8$\pm$2.6 & 43.5$\pm$2.1 & 67.5$\pm$0.6 & 53.5$\pm$0.9 & 67.5$\pm$2.7 & 55.8$\pm$1.9 \\
    GCN-{\em Muse} ($\mathbf{H}$) & 64.3$\pm$1.4 & 50.2$\pm$1.5 & 59.3$\pm$2.4 & 44.8$\pm$1.2 & 70.1$\pm$1.5 & 68.0$\pm$1.3 & 68.0$\pm$1.5 & 56.0$\pm$1.7 \\
    GCN-{\em Muse} ($\mathbf{U}$) & 64.9$\pm$1.6 & 51.4$\pm$1.8 & 60.5$\pm$2.0 & 46.2$\pm$2.2 & 70.8$\pm$2.3 & 67.5$\pm$2.0 & 68.5$\pm$1.7 & 56.1$\pm$1.9 \\
    GCN-{\em Muse} ($\mathbf{S^H}$) & 65.7$\pm$1.1 & 50.7$\pm$1.9 & 60.6$\pm$2.1 & 45.2$\pm$1.5 & 72.2$\pm$1.2 & 67.7$\pm$1.5 & 68.8$\pm$1.6 & 56.5$\pm$1.8 \\
    GCN-{\em Muse} ($\mathbf{S^U}$) & 65.9$\pm$0.9 & 52.0$\pm$2.0 & 61.4$\pm$1.8 & 45.7$\pm$2.0 & 72.0$\pm$1.8 & 67.2$\pm$2.3 & 69.5$\pm$2.0 & 57.4$\pm$1.4 \\
    GCN-{\em Muse} ($\mathbf{S^H+H}$) & 68.2$\pm$1.0 & 54.0$\pm$1.7 & 64.2$\pm$2.1 & 47.2$\pm$1.6 & 74.9$\pm$2.0 & 69.3$\pm$2.4 & 69.8$\pm$2.6 & 61.0$\pm$2.2 \\
    GCN-{\em Muse} ($\mathbf{S^U+U}$) & 67.6$\pm$1.3 & 55.0$\pm$1.5 & 65.3$\pm$2.3 & 47.8$\pm$2.2 & 74.2$\pm$2.3 & 70.5$\pm$2.0 & 70.0$\pm$2.3 & 60.5$\pm$2.0 \\
    GCN-{\em Muse} w/o $\mathcal{L}_p$ & 55.4$\pm$1.6 & 32.9$\pm$2.4 & 55.4$\pm$2.7 & 44.5$\pm$2.2 & 70.8$\pm$2.5 & 63.8$\pm$2.5 & 68.7$\pm$2.9 & 56.8$\pm$1.8 \\
    GCN-{\em Muse} w/o $MI$ & 66.2$\pm$1.0 & 57.1$\pm$2.0 & 64.0$\pm$2.5 & 47.9$\pm$2.1 & 75.4$\pm$2.4 & 70.0$\pm$2.8 & 71.1$\pm$2.5 & 60.8$\pm$2.0 \\
    2-GCN-{\em Muse} & 68.6$\pm$0.9 & 58.6$\pm$1.3 & 66.7$\pm$2.0 & \boldmath{}\textbf{48.9$\pm$1.9}\unboldmath{} & \boldmath{}\textbf{77.6$\pm$1.8}\unboldmath{} & 72.2$\pm$1.6 & 72.1$\pm$1.2 & \boldmath{}\textbf{62.6$\pm$2.3}\unboldmath{} \\
    GCN-{\em Muse} & \boldmath{}\textbf{69.5$\pm$0.7}\unboldmath{} & \boldmath{}\textbf{59.6$\pm$1.6}\unboldmath{} & \boldmath{}\textbf{67.4$\pm$2.4}\unboldmath{} & 48.7$\pm$2.0 & 77.4$\pm$2.3 & \boldmath{}\textbf{72.3$\pm$2.5}\unboldmath{} & \boldmath{}\textbf{72.6$\pm$2.8}\unboldmath{} & 62.0$\pm$1.7 \\
    \midrule
    GraphSAGE ($\mathcal{G}$) & 45.7$\pm$0.7 & 33.3$\pm$0.9 & 53.8$\pm$2.5 & 42.1$\pm$1.6 & 64.0$\pm$0.6 & 55.4$\pm$0.6 & 66.7$\pm$1.5 & 52.2$\pm$1.2 \\
    GraphSAGE ($\mathcal{G}^{'}$) & 44.9$\pm$0.8 & 31.7$\pm$1.2 & 50.2$\pm$2.9 & 39.5$\pm$1.9 & 64.1$\pm$0.8 & 54.0$\pm$0.4 & 66.1$\pm$1.6 & 51.8$\pm$1.3 \\
    GraphSAGE-{\em Muse} ($\mathbf{H}$) & 63.6$\pm$1.5 & 53.2$\pm$1.3 & 60.5$\pm$1.5 & 43.5$\pm$1.5 & 69.6$\pm$1.4 & 67.2$\pm$1.4 & 68.1$\pm$1.3 & 54.2$\pm$1.8 \\
    GraphSAGE-{\em Muse} ($\mathbf{U}$) & 64.6$\pm$1.1 & 52.7$\pm$1.5 & 62.6$\pm$2.1 & 44.0$\pm$1.6 & 69.1$\pm$1.6 & 65.9$\pm$2.4 & 67.1$\pm$2.0 & 54.5$\pm$1.5 \\
    GraphSAGE-{\em Muse} ($\mathbf{S^H}$) & 64.3$\pm$1.0 & 53.7$\pm$1.3 & 61.8$\pm$1.6 & 44.4$\pm$1.8 & 71.3$\pm$1.4 & 68.1$\pm$1.1 & 68.2$\pm$1.6 & 55.8$\pm$1.5 \\
    GraphSAGE-{\em Muse} ($\mathbf{S^U}$) & 65.1$\pm$0.8 & 53.6$\pm$1.2 & 62.2$\pm$1.9 & 43.7$\pm$1.5 & 70.5$\pm$2.0 & 67.0$\pm$2.0 & 66.7$\pm$2.0 & 55.2$\pm$1.7 \\
    GraphSAGE-{\em Muse} ($\mathbf{S^H+H}$) & 67.1$\pm$1.1 & 57.0$\pm$1.2 & 63.3$\pm$2.0 & 45.8$\pm$2.0 & 74.0$\pm$2.4 & 68.9$\pm$2.2 & 68.8$\pm$1.9 & 57.9$\pm$2.1 \\
    GraphSAGE-{\em Muse} ($\mathbf{S^U+U}$) & 66.5$\pm$0.9 & 56.5$\pm$1.0 & 64.7$\pm$1.7 & 45.3$\pm$1.3 & 73.5$\pm$1.6 & 68.5$\pm$1.9 & 69.0$\pm$2.2 & 58.0$\pm$2.4 \\
    GraphSAGE-{\em Muse} w/o $\mathcal{L}_p$ & 53.3$\pm$1.3 & 44.8$\pm$1.8 & 54.2$\pm$1.9 & 43.0$\pm$1.4 & 66.8$\pm$2.8 & 60.1$\pm$2.3 & 67.0$\pm$1.7 & 55.8$\pm$1.5 \\
    GraphSAGE-{\em Muse} w/o $MI$ & 64.9$\pm$1.1 & 56.0$\pm$1.5 & 63.0$\pm$2.0 & 45.8$\pm$1.8 & 72.4$\pm$2.5 & 70.6$\pm$1.8 & 70.2$\pm$1.9 & 58.8$\pm$1.7 \\
    2-GraphSAGE-{\em Muse} & \boldmath{}\textbf{68.4$\pm$1.1}\unboldmath{} & 57.7$\pm$0.9 & 66.2$\pm$2.6 & 48.0$\pm$1.7 & 75.4$\pm$2.2 & 70.7$\pm$1.3 & 70.4$\pm$1.8 & 60.3$\pm$1.4 \\
    GraphSAGE-{\em Muse} & 68.0$\pm$0.8 & \boldmath{}\textbf{58.6$\pm$1.7}\unboldmath{} & \boldmath{}\textbf{66.9$\pm$2.7}\unboldmath{} & \boldmath{}\textbf{48.3$\pm$1.7}\unboldmath{} & \boldmath{}\textbf{75.9$\pm$2.5}\unboldmath{} & \boldmath{}\textbf{71.0$\pm$2.2}\unboldmath{} & \boldmath{}\textbf{70.6$\pm$2.1}\unboldmath{} & \boldmath{}\textbf{60.8$\pm$2.0}\unboldmath{} \\
    \bottomrule
    \end{tabular}%
    }
  \label{tab:ablation}%
\end{table*}%

To investigate the effectiveness of fusing two views of subgraphs (\textbf{Q2}), we conduct the following three ablation studies.


(1) We decompose the proposed {\em Muse} into two components, \emph{i.e.}, GCN ($\mathcal{G}$) and GCN ($\mathcal{G}^{'}$) so as to verify whether the proposed {\em Muse} can capture distinct topological information. GCN ($\mathcal{G}$) only has the top branch in Fig. \ref{fig:pipeline} which can only learn representations from the raw graph, while GCN ($\mathcal{G}^{'}$) only has the bottom branch in Fig. \ref{fig:pipeline}. As can be seen from Table \ref{tab:ablation}, compared with {\em Muse}, only extracting representations from $\mathcal{G}$ and $\mathcal{G}^{'}$ has a sharp drop of up to $20\%$ in accuracy performance. Therefore, this paper designs two branches to learn embeddings under the latent and raw graph structure respectively, thus the merits of both topological structures are kept.

(2) Furthermore, we analyze how different representations, \emph{i.e.}, different node embedding and different subgraph embedding, affect the classification performance. In our work, four types of embedding (\emph{i.e.}, naive subgraph $\widetilde{\mathbf{S}}^{\mathbf{H}}$, latent subgraph $\widetilde{\mathbf{S}}^{\mathbf{U}}$, latent embedding $\mathbf{U}$, and naive embedding $\mathbf{H}$) are concatenated together for calculating the classification loss $\mathcal{L}_c$. 
To investigate the quality of each embedding, we design 6 versions of {\em Muse} by using different embeddings. As shown in Table \ref{tab:ablation}, $(\mathbf{H}), (\mathbf{U}), (\mathbf{S^H}), (\mathbf{S^U}), (\mathbf{S^H+H})$, and $(\mathbf{S^U+U})$ denote the model only using the corresponding embedding for calculating $\mathcal{L}_c$, \emph{e.g.}, $(\mathbf{S^U+U})$ is the model concatenating $\mathbf{S^U}$ and $\mathbf{U}$ as the augmented representation for calculating $\mathcal{L}_c$. From the experimental results, we can observe the models $(\mathbf{S^H})$ and $(\mathbf{S^U})$ are better than models ($\mathbf{H}$) and ($\mathbf{U}$), which indicates the subgraphs are more informative than nodes. Furthermore, by concatenating $\mathbf{S^H}$ and $\mathbf{H}$ (or $\mathbf{S^U}$ and $\mathbf{U}$), the performance is further improved. When concatenating $\mathbf{S^H}$, $\mathbf{H}$, $\mathbf{S^U}$, and $\mathbf{U}$, {\em Muse} can achieve the best performance. 
Besides, the above experimental results also confirm that different representations have different inductive biases. Therefore, it is necessary to design additional components, \emph{e.g.} $\mathcal{L}_p$, to alleviate the inductive biases.

(3)  We believe the complementary effects of these two streams are crucial. To verify this, we conduct ablation studies on naive and latent embeddings. We can observe that compared with only using one type of embeddings ({\em Muse}-$\mathbf{H}$ or {\em Muse}-$\mathbf{U}$), fusing two types of embeddings can significantly improve the accuracy. Besides, we create embeddings with two streams containing separate GNN models, and we can observe that Muse with two branches (2-GCN-{\em Muse} and 2-GraphSAGE-{\em Muse}) achieve results comparable to GCN-{\em Muse} and GraphSAGE-{\em Muse}, which indicates the information in the two streams are complementary.  However, the amount of parameters of {\em Muse} with two branches is almost twice that of {\em Muse}. 

To answer \textbf{Q3}, we study the effectiveness of each component by carrying out the following three ablation experiments.

(1) To verify the effectiveness of the prototypical loss $\mathcal{L}_p$, we compare {\em Muse} with {\em Muse} without $\mathcal{L}_p$.
In Table \ref{tab:ablation}, "w/o $\mathcal{L}_p$" means training the {\em Muse} without $\mathcal{L}_p$.
It can be obviously found that optimizing the prototypical loss $\mathcal{L}_p$ to leverage inductive bias is effective. By removing $\mathcal{L}_p$, the performance has a sharp drop of about $10\%$. The experimental results show that the issue of scarce labeled node settings can be further alleviated by considering inductive bias.

(2) We also investigate the effectiveness of optimizing subgraphs by maximizing mutual information $MI$. We randomly initialize the masks $\mathbf{M^H}$ and $\mathbf{M^U}$ without optimizing the mutual information $MI$. By comparing {\em Muse} with {\em Muse} w/o $MI$ in Table \ref{tab:ablation}, we find a performance degradation of about $3\%$. From this phenomenon, applying optimization for identifying subgraphs is necessary.

(3) To validate the generalization of {\em Muse}, we combine {\em Muse} with both spectral-based GNN, \emph{e.g.}, GCN~\cite{10GraphConvolutionalNetworks} and spatial-based GNN, \emph{e.g.}, GraphSAGE~\cite{10GraphSAGE}.  
As shown in Table \ref{tab:ablation}, GCN-{\em Muse} and GraphSAGE-{\em Muse} have made significant improvements over GCN and GraphSAGE.
As the label rates
get smaller, the improvement increases significantly. In particular, for one labeled sample per class, there
is 5\%$\sim$20\% improvement with our proposed framework.

To clearly visualize the effectiveness of the proposed {\em Muse}, we use t-SNE to visualize the embedding learned by GCN, GCN-{\em Muse}, GraphSAGE, and GraphSAGE-{\em Muse} on the Cora dataset. 
As shown in Fig. \ref{fig:tsne}, the embedding learned by GCN-{\em Muse} and GraphSAGE-{\em Muse} is more distinguishable and has compact clusters and clear boundaries.
The above analysis indicates the proposed {\em Muse} can achieve promising performance not only on spectral-based GNN but also on spatial-based GNN. 

\begin{figure*}[htbp] 
  \centering   
  \subcaptionbox{\zz{Averaged running time per epoch under the setting of 1 labeled node per class.}}{
  \includegraphics[width=3.8cm]{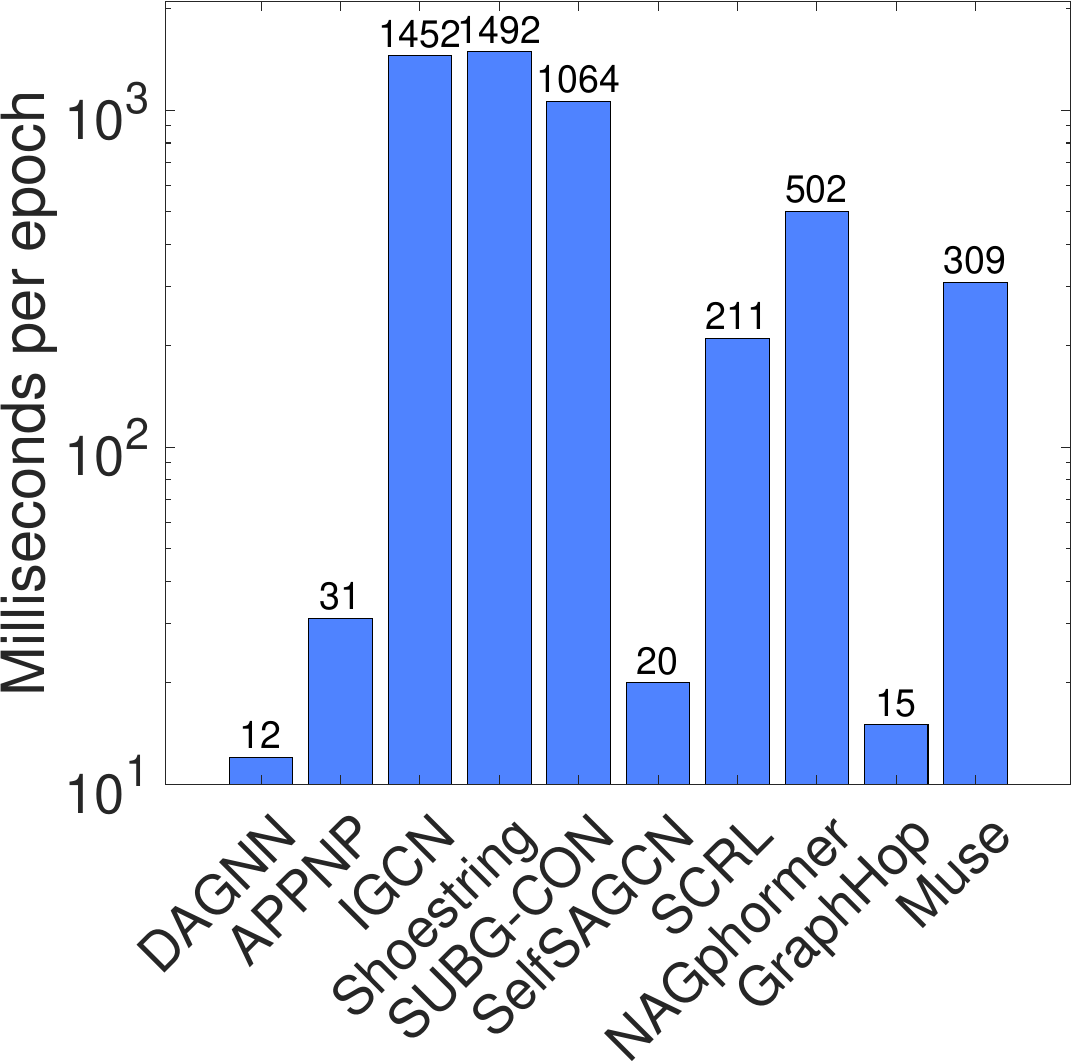}
  }
   \quad 
  \subcaptionbox{\zz{Total running time under the setting of 1 labeled node per class.}}{
  \includegraphics[width=3.9cm]{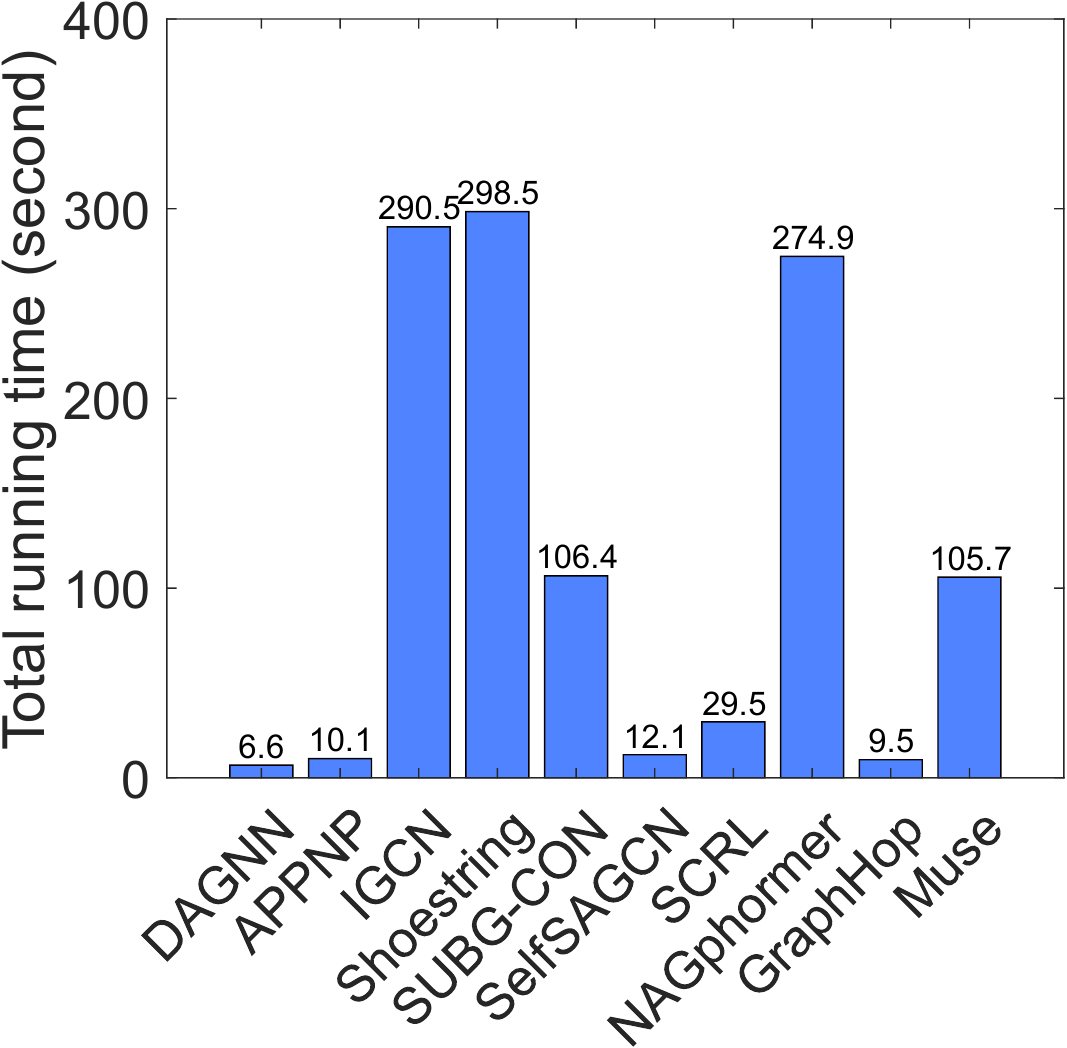}
  }
    \quad 
  \subcaptionbox{The correlation between the number of labeled nodes and computational cost.}{
  \includegraphics[width=3.8cm]{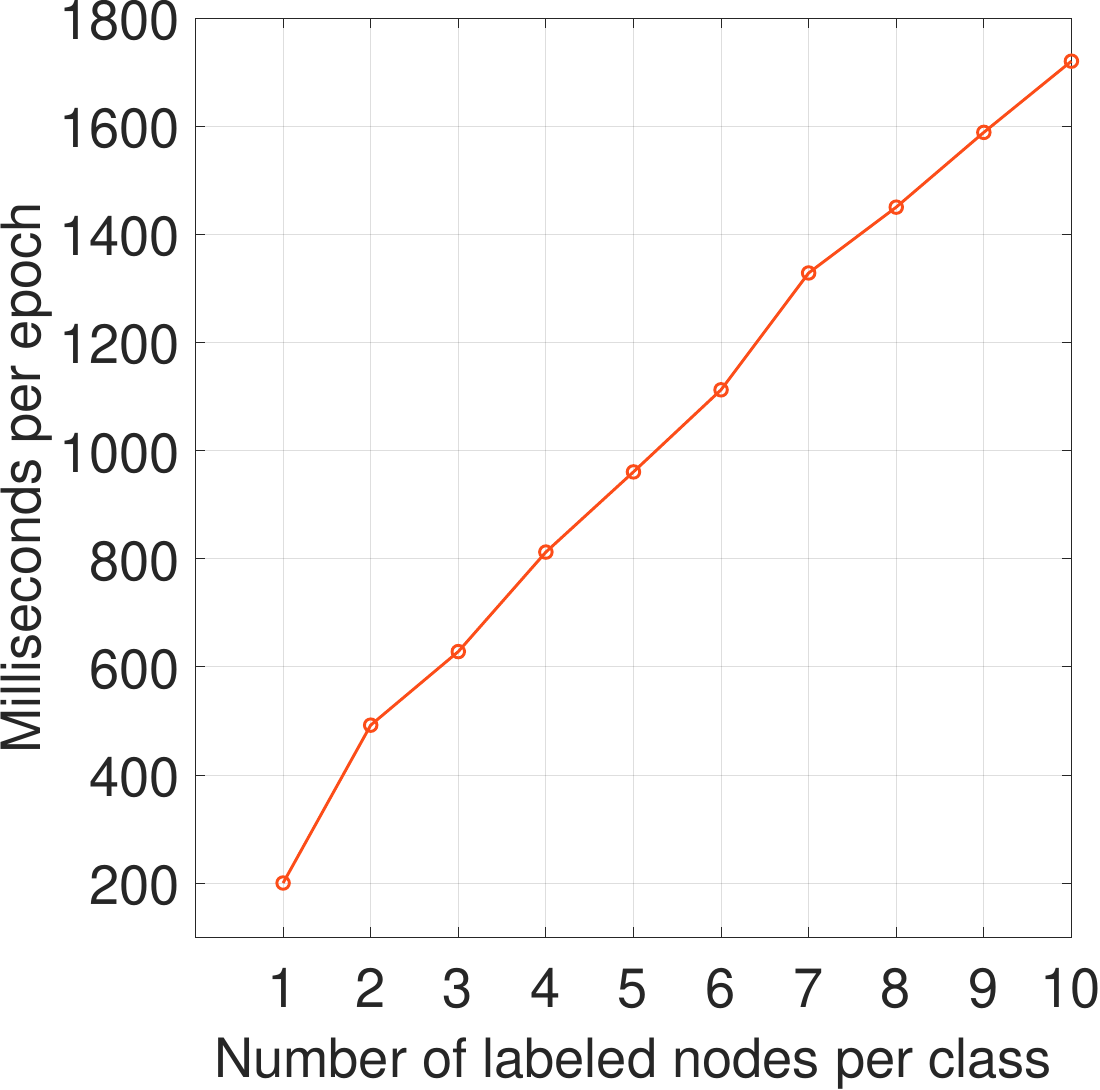}
  }
    \quad 
  \subcaptionbox{\zz{GPU memory usage under the setting of 1 labeled node per class.}}{
  \includegraphics[width=3.8cm]{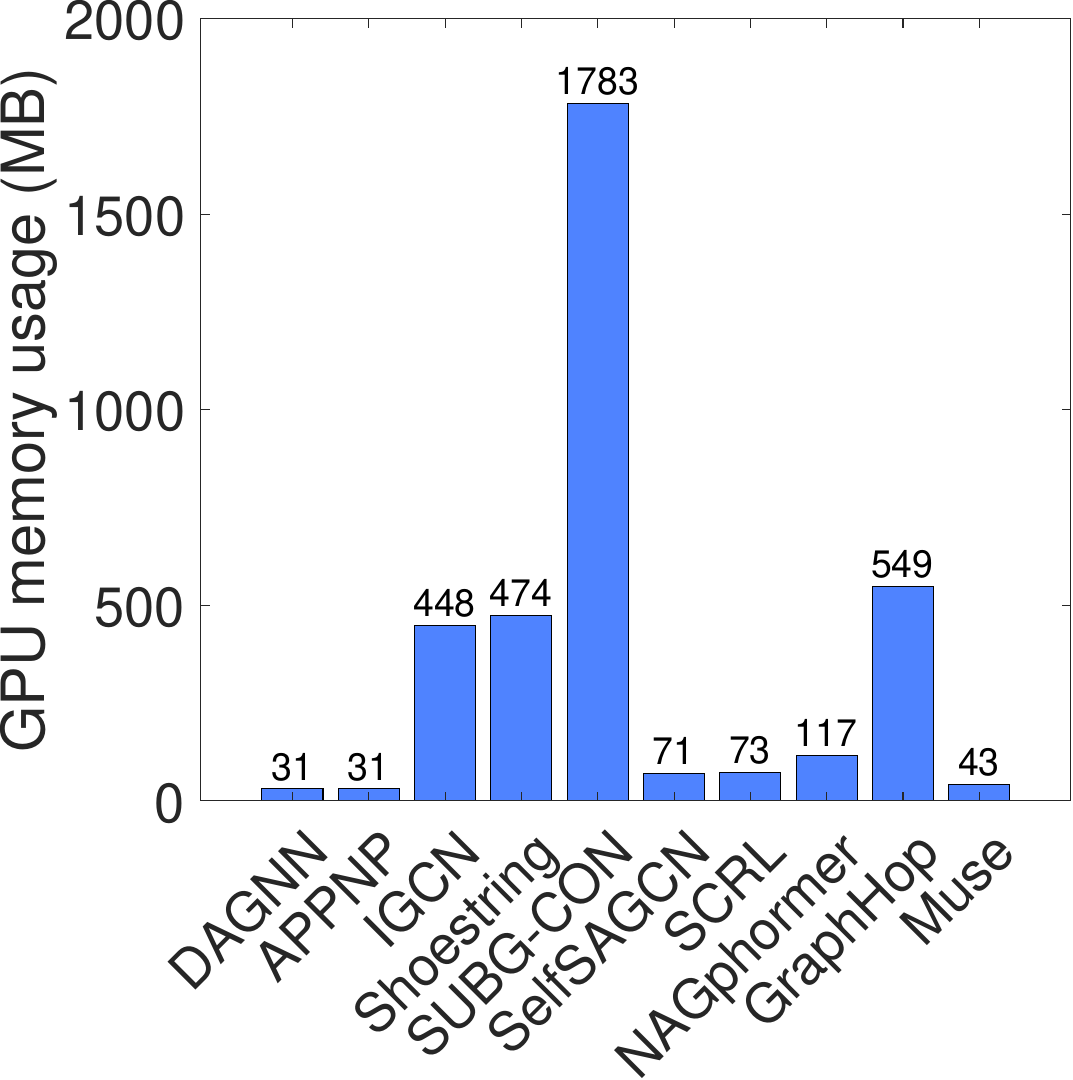}
  }
  \caption{The computational cost evaluation on Cora.}
  \label{fig:time}
\end{figure*} 



\subsection{Parameter Analysis}
To answer \textbf{Q4}, the behavior of the proposed {\em Muse} is analyzed on four key parameters: the parameter $\lambda_p$ for controlling the weight of the prototypical loss, the number of layers in GNN, the threshold values $k$ for masking the naive subgraph embedding, and the $\tau$ for masking the latent subgraph embedding.
In Fig. \ref{fig:parameter}, we can observe increasing $\lambda_p$ will improve the classification accuracy and converge around $\lambda_p$ of 4$\sim$5. The strong correlation between $\lambda_p$ and accuracy shows that $\mathcal{L}_p$ can play an effective role in preventing overfitting in scarce labeled node classification. For the number of layers in GNN, increasing the number of layers will reduce the accuracy due to over-smoothing.

The threshold values $k$ and $\tau$ of masks determine the scope of the local information and long-range dependencies that can be captured by subgraphs. As can be observed from Fig. \ref{fig:parameter}, increasing the parameter $k$ can not significantly improve the performance of the model, which means that long-range dependencies are not well captured by simply increasing the number of hops. While increasing the threshold $\tau$ can improve the accuracy and then converge around $\tau$ of 0.5. 

\subsection{Computational Cost Analysis}

The performance of {\em Muse} comes at a small price. We compare the mean training time cost per epoch of different algorithms including semi-supervised learning and SSL methods on the Cora dataset, as shown in Fig. \ref{fig:time} (a). \zz{In addition, we measure the total running time in seconds, as depicted in Fig. \ref{fig:time} (b), where the training epochs are set to 1000 with an early stopping criterion. It should be noted that the process of Isomap is included in the computational cost.} Despite the additional computational cost, {\em Muse} is still very competitive in terms of classification performance as compared to other algorithms.

To investigate the cost under different numbers of labeled nodes, we collect the training time under 1 to 10 labeled nodes on the Cora dataset. In Fig. \ref{fig:time} (c), we can find with the increase in the number of labeled nodes, the training time will increase linearly, as {\em Muse} optimizes two types of masks for each labeled node for identifying subgraphs for the node. \zz{For this reason, when there is a large amount of labeled data in the graph, directly applying the proposed {\em Muse} will result in a notable computational cost. The method is dedicatedly designed to the scenario where the labels are expensive to collect. In future work, we will explore efficient subgraph augmentation strategies for cases with abundant data availability.}

\zz{Furthermore, a statistical analysis of GPU memory usage during algorithm training was carried out, illustrated in Fig. \ref{fig:time} (d). Our proposed {\em Muse} exhibits notably low memory consumption, at just 43MB. It is worth noting that the memory usage of our algorithm does not increase with the number of labeled nodes, since we sequentially handle each labeled node in Algorithm 1.}


\section{Conclusion}

In this work, we advanced graph-based SSL when labeled data are severely scarce. Specifically, we propose a multi-view subgraph neural network ({\em Muse}) that can handle the long-range dependencies of nodes. In this process, two views of subgraphs are identified from the input data space and the latent space for augmenting the supervision signals. By fusing these views of subgraphs, our proposed {\em Muse} can capture not only local structure information but also correlated, distant, yet informative information from a large number of unlabeled nodes. We show that the generalization capability of our model can be further boosted with various inductive biases. In addition, the experiments on canonical node classification tasks with graph data exhibit substantial improvements compared with alternative baselines. 
Several possible directions may be taken for future work. For example, the proposed {\em Muse} could be extended to solve graph classification tasks, \emph{e.g.,} molecular property prediction. In addition, a rich body of information theory methods can be explored to capture the most informative knowledge among the nodes.

.

\bibliographystyle{IEEEtran}

\end{document}